\documentclass[journal]{IEEEtran}
\usepackage{amsmath,amsfonts}
\usepackage{algorithmic}
\usepackage{algorithm}
\usepackage{array}
\usepackage[caption=false,font=normalsize,labelfont=sf,textfont=sf]{subfig}
\usepackage{textcomp}
\usepackage{url}
\usepackage{verbatim}
\usepackage{graphicx}
\usepackage{cite}
\usepackage{stfloats}
\usepackage{booktabs}
\usepackage{color}
\usepackage[dvipsnames]{xcolor}
\usepackage{dsfont}
\usepackage{colortbl}
\usepackage{hyperref}
\hypersetup{
    colorlinks = true,
    linkcolor = blue,
    citecolor = blue,
    urlcolor = blue
}
\usepackage{multirow}
\usepackage{soul}
\usepackage{ragged2e}

\usepackage{multicol}
\usepackage{makecell}
\usepackage{booktabs}
\usepackage{amssymb}
\usepackage{bbding}
\usepackage{pifont}
\usepackage{wasysym}

\soulregister\cite7
\soulregister\citep7
\soulregister\citet7
\soulregister\ref7
\soulregister\eqref7
\begin{document}

\title{Multi-Agent Reinforcement Learning for Autonomous Driving: A Survey}

\author{Ruiqi Zhang$^{1, 2}$,
        Jing Hou$^{1}$,
        Florian Walter$^{3}$,~\IEEEmembership{Member,~IEEE},
        Shangding Gu$^{4}$, 
        Jiayi Guan$^{1}$,
        Florian Röhrbein$^{5}$,\\~\IEEEmembership{Senior Member,~IEEE},
        Yali Du$^{6}$,
        Panpan Cai$^{7}$,
        ~Guang Chen$^{1, 4,*}$,~\IEEEmembership{Member,~IEEE},
        and Alois Knoll$^{4}$,~\IEEEmembership{Fellow,~IEEE}
\thanks{
$^{1}$Ruiqi Zhang, Jing Hou, Jiayi Guan and Guang Chen are with Tongji University, Shanghai, China. 
$^{2}$Ruiqi Zhang is with University of California, Berkeley, United States.
$^{3}$Florian Walter is with Machine Intelligence Lab, University of Technology Nuremberg, Germany.
$^{4}$Shangding Gu, Guang Chen and Alois Knoll are with Technical University of Munich, Germany.
$^{5}$Florian Röhrbein is with Chemnitz University of Technology, Germany.
$^{6}$Yali Du is with King's College London, United Kingdom.
$^{7}$Panpan Cai is with Shanghai Jiao Tong University, China.
$^{*}$Corresponding author: Guang Chen (mail to: \tt{guang@in.tum.de})
}
}

\markboth{IEEE Journals, PrePrint Verision}%
{Shell \MakeLowercase{\textit{et al.}}: A Sample Article Using IEEEtran.cls for IEEE Journals}

\maketitle

\begin{abstract}
Reinforcement Learning~(RL) is a potent tool for sequential decision-making and has achieved performance surpassing human capabilities across many challenging real-world tasks. As the extension of RL in the multi-agent system domain, multi-agent RL~(MARL) not only need to learn the control policy but also requires consideration regarding interactions with all other agents in the environment, mutual influences among different system components, and the distribution of computational resources. This augments the complexity of algorithmic design and poses higher requirements on computational resources. Simultaneously, simulators are crucial to obtain realistic data, which is the fundamentals of RL. In this paper, we first propose a series of metrics of simulators and summarize the features of existing benchmarks. Second, to ease comprehension, we recall the foundational knowledge and then synthesize the recently advanced studies of MARL-related autonomous driving and intelligent transportation systems. Specifically, we examine their environmental modeling, state representation, perception units, and algorithm design. Conclusively, we discuss open challenges as well as prospects and opportunities. We hope this paper can help the researchers integrate MARL technologies and trigger more insightful ideas toward the intelligent and autonomous driving.
\justifying
\end{abstract}



\section{Introduction}
\IEEEPARstart{L}{arge-scale} 
autonomous driving systems have attracted tons of attention and millions of funding from industry, academia, and government in recent years\cite{TCyberDRL4MAS, TIV2023survey}. The motivation behind developing such a system is to replace human drivers with automated controllers. It can significantly reduce the time consumption and workload of driving, enhance the efficiency and safety of transportation systems, and promote economic development. Generally, to detect the vehicle states and generate reliable control policies, automated vehicles~(AVs) should be equipped with massive electric units, like visual sensors including radars, light detection and ranging~(LiDAR), RGB-Depth~(RGB-D) cameras, event cameras, inertial measurement units ~(IMU), global positioning system~(GPS) and so on\cite{zhang2014loam, neurodfd, eventsurvey}. A salient challenge in this topic is to build a robust and efficient algorithm that is capable of processing massive information and translating this data into real-time operations. Early works divide this big issue into perception, planning, and control problems and solve them independently, known as modular autonomous driving. 

On the other hand, as a powerful toolkit for sequential decision-making, reinforcement learning~(RL) can optimize agent behavior models with the reward signal. As its evolution, deep RL combines the advantages of RL and deep neural networks, which enable to abstract complex observations and learn efficient feature representations\cite{sutton2018reinforcement}. In the past representative research, it has exhibited performances in domains such as board games\cite{silver2018general, silver2017mastering}, video games\cite{vinyals2019grandmaster, fuchs2021super} and robotic control\cite{ruiqi2022residual, song2023reaching, kaufmann2023champion}, where it rivaled or even surpassed human performances. For autonomous driving, RL brings end-to-end control into reality, which transitions directly from what the vehicle senses to what the vehicle should do, like human drivers. While RL has obtained many remarkable achievements on AVs, most of the related work has approached the issue from the perspective of individual vehicles, which leads to self-centric and possibly aggressive driving strategies, which may cause safety accidents and reduce the efficiency of transportation systems.

\begin{figure}[tbp]
    \centerline{
    \includegraphics[width=9cm]{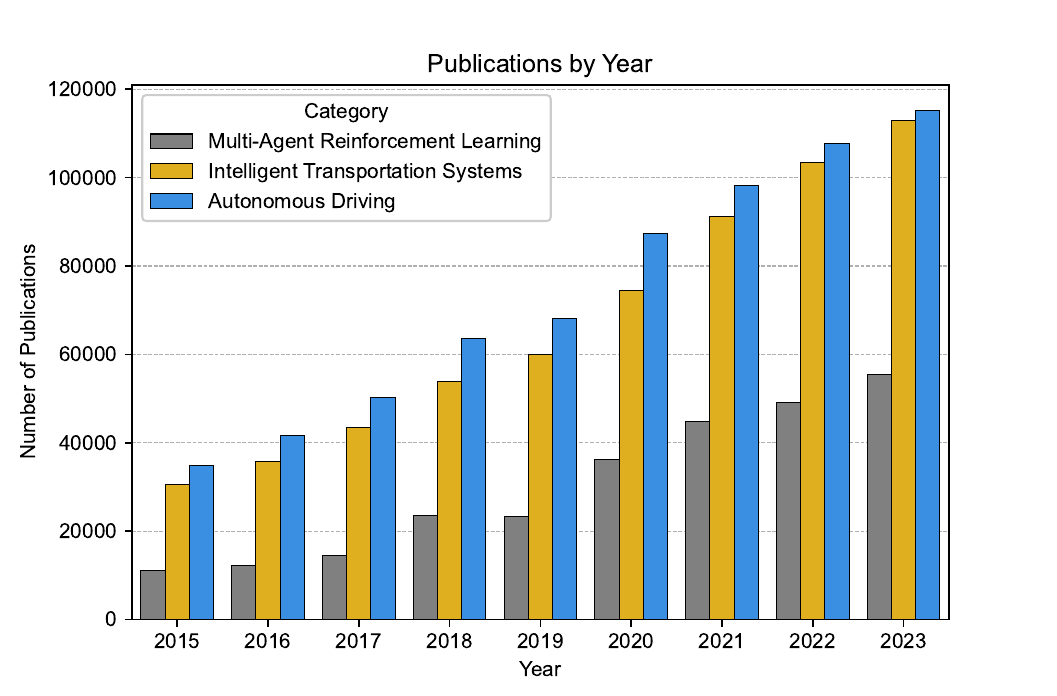}
    }
    \caption{The number of keywords \textit{multi-agent reinforcement learning}, \textit{autonomous driving} and \textit{intelligent transportation systems} publications from 2015 to 2023 (from Dimension AI\cite{dimensions}). These three research topics are in rapid development and obtaining increasing attention from academia.}
    \label{trend}
    \vspace{-10pt}
\end{figure}

For real-world traffic systems, we typically define them as multi-agent systems (MAS) and aim to optimize the efficiency of the entire system rather than merely maximizing individual interests. In MAS, all agents make decisions and interact within a shared environment. This means that the states of each agent depend not only on its actions but also on the behaviors of others, making the environmental dynamics non-stationary and time-variant. Additionally, depending on the task setup, agents may either cooperate or compete with each other. In such complex scenarios, manually programming a priori actions is nearly impossible\cite{gronauer2022marl2}.
Thanks to significant advancements in multi-agent reinforcement learning (MARL), substantial breakthroughs have been achieved in traffic control\cite{traffic1, traffic2}, energy distribution\cite{perolat2017energy1, energy2}, large-scale robotic control\cite{copo, ruiqi2022pipo}, and economic modeling and prediction\cite{trade1, trade2}. Figure 1 illustrates the number of publications on these related research topics. Using the Dimensions database AI search\cite{dimensions}, we searched for keywords including \textit{multi-agent reinforcement learning}, \textit{autonomous driving}, and \textit{intelligent transportation}. The statistical results show that academia is highly interested in these issues, and the related research field is in a rapid growth phase.
To accelerate further research and assist new researchers in getting started quickly, we have reviewed over 200 publications, open-source software, and code bases, and then systematically summarized existing achievements and the latest advancements in this paper.

\begin{figure*}[tbp]
    \centerline{
    \includegraphics[width=18cm]{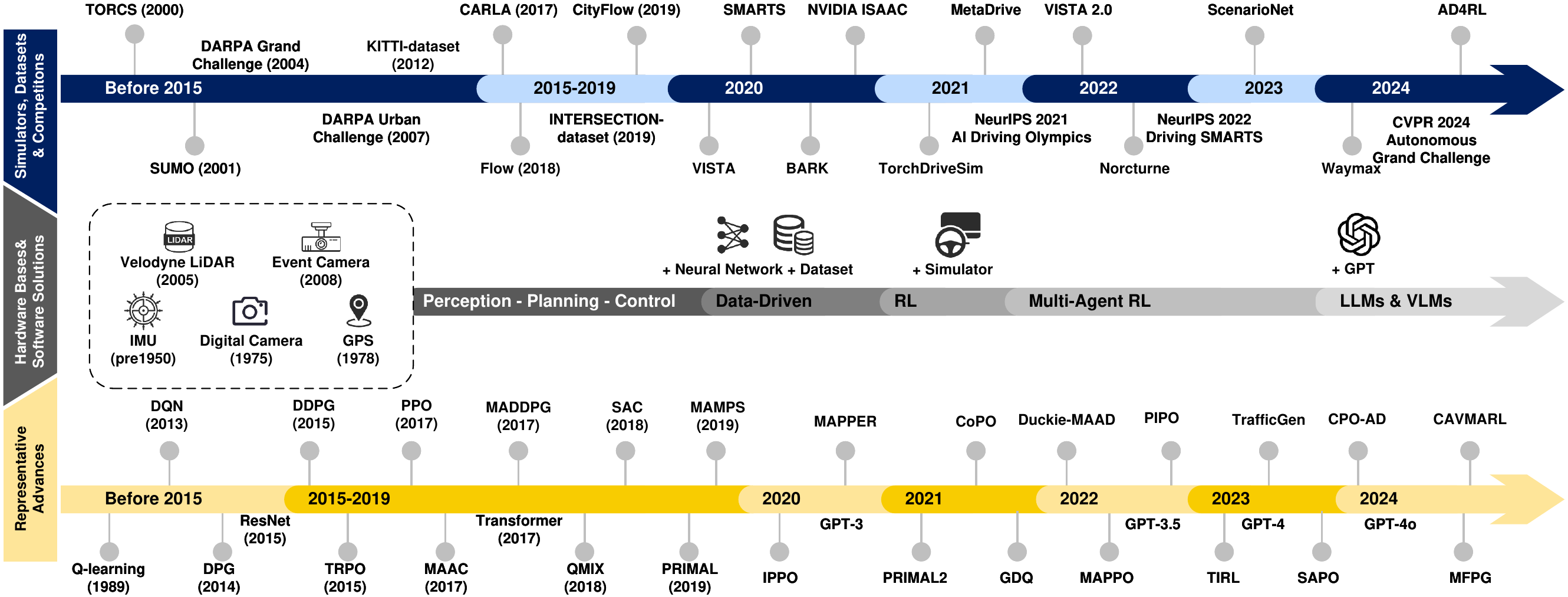}
    }
    \caption{Timeline of the evolution and representative studies of autonomous driving, deep learning and RL. Based on existing hardware, early research was conducted with a hierarchical scheme, i.e. perception-planning-control. Around 2014, with the rapid development of deep learning and the emergence of datasets, data-driven methods became the mainstream for a time. From 2015 to 2019, numerous RL algorithms appeared, and people realized the opportunities of end-to-end control through simulators. In 2017, MADDPG\cite{maddpg2017} introduced single-agent RL into multi-agent systems, leading to massive subsequent research on how to control large-scale autonomous vehicles through MARL. In 2020, ChatGPT was launched and made language models receive massive attention. }
    \label{fig::history}
    \vspace{-5pt}
\end{figure*}
Here, we note other recent reviews. In the milestone series\cite{milestonesos, milestone1, milestone2}, the authors summarized the blueprint from the history to the future and introduce the influential algorithms in autonomous driving briefly. There are also many surveys\cite{Aradi2020survey, kiran2021deep, ailab2023e2esurvey} that introduce the basic theory and application of RL and analyze the state-of-the-art~(SoTA) algorithms for autonomous driving at their published time, but they mainly focus on single-agent learning. Authors of survey\cite{TIV2016survey} first defined the hierarchical-structured autonomous driving system and limited their scope to local motion planning. They illustrated the kinetics of vehicle and demonstrated how sampling and search-based approaches work mathematically. However, they ignored the contribution of learning-based methods. In the recent survey of motion planning\cite{TIV2023survey}, researchers comprehensively investigated the pipeline and learning methods including deep learning, inverse RL and imitation learning, and MARL. Similarly, a detailed overview covers the latest taxonomy and methodology in trajectory prediction\cite{TIV2022survey}. There are some excellent reviews that summarized MARL approaches for AVs\cite{TCyberDRL4MAS, itscmarlsurvey, krdsurvey}. Nevertheless, researchers have made significant progress in theory and application, as well as in advanced robotic simulators in recent years. As a crucial component of online RL training, simulators determine the sim-to-real gap, that is, whether the policy learned by the agents can be easily transferred to physical robots. Consequently, to enable engineers and researchers to capture the latest updates and accelerate technological progress, we comprehensively summarize the technologies, challenges, and prospects in this field. Generally, the main contribution of this paper can be summarized as follows:
\begin{itemize}
    \item We propose a series of criteria of benchmarks, analyzing and summarizing the features of advanced simulators, datasets and competitions of large-scale autonomous driving in detail.
    
    \item We categorize the state-of-the-art MARL methodologies, and review their technical improvements, insights and unsolved challenges in this field comprehensively.
    
    \item We capture the latest advances from relevant fields and delve into future directions of MARL-based autonomous driving from multiple perspectives and dimensions. 
    
    \item We publish and maintain the GitHub repository\footnote{[Online]. Available: \url{https://github.com/ispc-lab/MARL4AD}} to continuously report and update the latest studies in MARL-based autonomous driving, intelligent transportation systems and other relevant areas.
\end{itemize}

In Fig.~\ref{fig::history}, we visualize the history of development in MARL, datasets, simulators, hardware and software for autonomous driving, and other related fields. In general, with the development of large-scale datasets and deep learning, autonomous driving has stepped from hierarchical control to the data-driven era. With the emergence of advanced simulators, the RL-based method steps onto the stage, and then new techniques like large language models bring more opportunities. We will analyze them later in detail, and the rest of this article is organized as follows: In Section~\ref{simu}, we first describe the metrics of benchmarks. We also analyze features of the most advanced autonomous driving simulators and datasets. In Section~\ref{prelim}, we recall the basic concepts, definitions, and open issues in RL and MARL. In Section~\ref{method}, we exhaustively introduce the SoTA MARL algorithms for autonomous driving. Specifically, we analyze their state and action setup, methodological insights and applications. In Section~\ref{challenge}, we point out the existing challenges and give out the possible solutions. In Section~\ref{future}, we capture the latest advances and propose promising directions toward safer and more intelligent autonomous driving. 


\section{Autonomous Driving Benchmarks}
\label{simu}

RL is always data-hungry. Generally, it requires continuous interaction with the environment to obtain behavior trajectories, which facilitates more accurate value estimations from deep neural networks~\cite {drlsurvey, drlsurvey2}. However,  due to the economic damage caused by uncertain exploration processes, we typically would not deploy our RL policies on real robots directly. Consequently, within the RL paradigm, data from real driving and high-fidelity simulators are ubiquitously adopted in the development of RL-based autonomous driving. In this section, we will introduce various data sources for large-scale MARL in autonomous driving and traffic systems.

\subsection{What is important for a good benchmark?}
A benchmark involves the simulation of physical models, optical rendering, environment and interaction mechanisms, algorithms, and other complex tasks. For MARL-based autonomous driving, we identify the following crucial criteria. We list them out here and will analyse existing data resourses by these metrics.

\subsubsection{Realism and Fidelity} 
High realism and fidelity ensure that the simulator can accurately replicate real-world driving conditions like weathers, lights, environmental dynamics, etc., and mitigate their distributional bias before real-world deployment. Deep learning models especially deep RL demand accurate data. In this case, the realism simulator ensures that its data is representative of real-world scenarios.

\subsubsection{Scalability}  
Scalability ensures that simulators can handle the dynamic environments with numerous entities and variables, so that we can mimick the complexity of real-world scenarios. Scalable MARL algorithms can efficiently learn from the interaction among time-variant numbers of agents like the real traffic system.

\subsubsection{Diversity} 
Diverse scenarios ensure that vehicles can be tested in a wide range of situations, including different traffic conditions, weather, and road structures. For deployment in the real world, it also contributes to the development of more robust autonomous driving systems and makes AVs more reliable in unpredictable real-world conditions. Simultaneously, diversity also implies that the agents could be heterogeneous, which brings more complex interactions and matches to the realities of actual traffic systems.

\subsubsection{Efficiency} 
Time and computational resources are both significant concerns for MARL and autonomous driving\cite{ViewEfficiency,hou2023spreeze}. Lightweight simulators reduce computational consumption, and experiments work on cheaper and smaller hardware. Meanwhile, highly-parallelized simulator allows multiple environments to run concurrently and promotes the training process of MARL algorithms. Note that there is always a hard trade-off between fidelity and efficiency\cite{ViewTradeoff}. High fidelity requires complex computations to replicate real-world scenarios, especially for 3D visual information.

\subsubsection{Transferability} 
Transferability requires the simulator to support various sensors technically and to replicate their characteristics. For autonomous driving tasks, there are significant differences in the data formats and frame rates of LiDAR, IMU, and cameras. It is essential for the simulator to maintain consistency in the sensor parameters of the intelligent agents with those of commercially available devices. Moreover, transferability is also presented in the simulator's compatibility with the vehicle's device in terms of programming language, communication protocols, and computing platforms.

\subsubsection{Features, Maintenance and Supports} 
Reproducing the effectiveness of algorithm is always time-consuming. Therefore, it would be beneficial for developers to provide fundamental and verified baselines for testing, also with user-friendly application programming interfaces~(APIs), annotations and tutorial documentation. These provisions would establish a fair and open comparison standard and improve the efficiency of subsequent developments. Furthermore, lasting maintenance is necessary. Hardware and software frequently discard old features and develop new ones, which can make data and code outdated. Therefore, continuous maintenance of datasets and code bases is an important task.

\subsection{Advanced Simulators}
The selection of a simulator for MARL-based autonomous driving is a critical step. A good selection would save resources and  enable the generation of vast amounts of diverse and high-quality training data, which is essential for effectively training MARL algorithms. Additionally, the simulator allows for parallelized testing and training and significantly accelerates the development process by reducing the time required to experiment with various scenarios and conditions. The advantages on extensive data generation and enhanced time efficiency make simulators indispensable for advancing autonomous driving technologies through MARL.

\subsubsection{The Open Racing Car Simulator}
\textit{TORCS}\cite{torcs} was first released in 2000. As a highly modular simulator for multi-agent racing, each race car offers low-level APIs to access partial vehicle states and provides visual information from multiple perspectives. After decades of evolution, it has accurate and editable vehicle dynamics, including the rotational inertia of different components, mechanical structures, tire dynamics, and a simplified aerodynamic model. The simulator supports discrete-time simulations with high frequency up to 500Hz, which allows for the development of complex, high-speed, and aggressive driving controllers on it\cite{wymann2015torcs}. Many representative competitions and RL-based research conduct their research on TORCS\cite{SCR, torcsresearch, torcsRLtutorial}. Afterward, \textit{Gym-TORCS} was released and aligned to OpenAI Gym\cite{openaigym}, which provides unified APIs and a Python wrapper to facilitate the rapid development of RL-based controller\cite{gymtorcs}. However, it still lacks support for MARL and a paralleled environment. Later, \textit{MADRaS}\cite{santara2021madras} filled this vacancy and offered both single-agent and multi-agent environments and interfaces, which could be used to test autonomous vehicle algorithms both heuristic and learning based on an inherently multi-agent setting.

\subsubsection{Simulation of Urban Mobility} 
\textit{SUMO}\cite{SUMO} has become a famous benchmark for the simulation of large road networks. It supports a wide range of scenarios and rich APIs so that users can customize traffic scenarios easily. Meanwhile, SUMO provides well-documented tutorials to assist users in implementing their simulations. In recent years, numerous studies have utilized this simulator to develop efficient and safe MARL algorithms for complex scenarios and obtain notable achievements\cite{sumoR1, sumoR2, sumoR3}. To accelerate RL research, developers released a new simulator \textit{Flow}\cite{Flowsimulator} with interfaces to the distributed RL framework RLLIB\cite{liang2018rllib} to achieve high-frequency traffic flow simulation. At the same time, it permits the integration with Amazon Web Services (AWS) elastic compute cloud and expands the variety of controllers, which brings higher flexibility and enables the training of large-scale RL policies. In \textit{CityFlow}\cite{zhang2019cityflow}, developers improved their computational speed to 20 times higher than SUMO. Hence, the real-time simulation of city-level traffic networks becomes possible. CityFlow also expands interfaces for MARL algorithms and allows external data import, which means it can simulate accurate data and generate nearly authentic samples for policy optimization.

\subsubsection{Scalable Multi-Agent Reinforcement Learning Training School}
\textit{SMARTS}\cite{zhou2020smarts} is one of the most advanced traffic simulators proposed by Huawei Noah's Ark Lab. It is established on the Social Agent Zoo platform and provides various heterogeneous agent assets for structured traffic flow. SMARTS also has Gym-standardized APIs and integrates broader MARL libraries like PyMARL\cite{pymarl}, MALib\cite{malib} and RLLIB. Moreover, it supports SUMO as the background provider but optimizes its vehicle dynamics with a Bullet-based physical engine\cite{bullet}. Implementing high-speed distributed computation introduces a bubble mechanism, which allows the elastic assignment of computational resources on local or remote machines. Furthermore, it offers a strong visualization toolkit through web streaming and allows developers to monitor the process from anywhere. Unlike other one-off works, SMARTS has established a substantial and stable community and promotes many promising works\cite{guantvt, smartR2, smartR3, sapo}. So far, its developers have maintained and constantly expanded the simulator's functionalities. 

\subsubsection{MetaDrive}
\textit{MetaDrive}\cite{metadrive} is one of the latest multi-agent system simulators based on Panda3D\cite{panda3D}, possessing a broad asset library and allowing the import of external data. Beyond the given structured scenarios, developers can easily customize road map and traffic flows, and set the attributes of scene components via high-level APIs. Meanwhile, it establishes hierarchical management of assets by defining four types of manager classes, which facilitates developers in customizing agent mixtures, interactions and policy generalizability tests. Theoretically, MetaDrive can create an infinite variety of traffic scenarios. It employs state vectors as the agents' observations and provides abundant RL benchmarks, including model-free RL\cite{ppo, sac}, imitation learning\cite{imitationl}, and offline RL\cite{offlinerl}. Although it forsakes fine-grained visual information, it enables more rapid and efficient simulation and has triggered many insightful works\cite{copo, lan2023trafficgen}.

\subsubsection{CAR Learning to Act}
\textit{CARLA}\cite{carla} is one of the state-of-the-art open-source 3D simulators for its realistic Unreal Engine 4-based dynamics simulation, lightweight optical rendering, and comprehensive technical support. For environmental information, it provides detailed scenes with various architectures, road configurations, and natural conditions, especially diverse weather settings, which are beneficial for the generalizability test of policy. Through its free asset library, it can simulate high-density traffic flow, pedestrians with different behaviors and traffic lights and signs, which makes it possible for automated agents to comprehend traffic regulations. Significantly, CARLA supports various sensors including LiDAR, RGB-D cameras, GPS, radar, and event cameras with editable characteristics and realistic noise. With rich C++ and Python-based APIs, researchers can freely define the attributes of agents and then analyze the collected data. Nowadays, CARLA not only makes substantial contributions to the MARL field but also continuously impacts computer vision research. A welcome trend is that its developers have established an official website with a good tutorial, blog, and user community for further updates and improvement. As a good supplement, based on CARLA, some researchers have developed a new benchmark named \textit{MACAD}\cite{macad2020} with a faster implementation for MARL and integrated Gym-like APIs. Generally, CARLA significantly facilitates systematic research\cite{carlaR1, carlaR2, carlaR3} for vision-based driving and reduces the simulation-to-reality~(sim-to-real) gap, which is crucial for deploying the large-scale driving controller.

\subsubsection{Virtual Image Synthesis and Transformation for Autonomy}
\textit{VISTA}\cite{vista1} is a data-driven simulator and synthesizes time series of perceptual inputs from real-world. In contrast to physical simulators, VISTA aims to reconstruct that world and synthesize novel viewpoints within the environment via inputting real data of the physical world. Different sensing modalities, environments, dynamics, and tasks with varying complexity are supported.Meanwhile, it is highly modular, customizable, and extensible. Since the behavior trajectories are generated from real data, the sim-to-real gap is minimized, which is empirically validated by researchers in real-world experiments. In the subsequent \textit{VISTA 2.0}\cite{vista2}, researchers integrate UNet architecture\cite{unet} to reproduce the dense output of LiDAR and apply temporal interpolation to estimate the events through RGB images, providing an additional two sensor types and data formats for AVs. Simultaneously, researchers also introduce a version that supports multi-agent interactions and validates basic scenarios involving multiple AVs\cite{vista2MARL}. However, its application in large-scale complex scenarios remains unexplored. 


\subsubsection{NVIDIA ISAAC-Sim}
\textit{ISAAC-Sim}\cite{isaacsim} is established on the PhysX5 engine and supports GPU-based photorealism with real-time ray tracing and the MDL material definition for physically based rendering, which provides nearly real scenarios for robotic training and testing. Besides its rich asset library, it also permits developers to import mesh models, customized assets, and URDF format robot structure files, which significantly enhance the flexibility in robotic research. More importantly, it facilitates communication with other important tools like ROS\cite{ros1}, ROS2\cite{ros2}, and Gazebo\cite{gazebo}, simplifying the real-world deployment. In the latest version, human simulation is introduced and supports the simulation for available LiDAR brands like Ouster, Hesai, and Slamtec. Ray-traced~(RTX) technology, can feedback more accurate LiDAR data w.r.t. various reflective materials or under different conditions.  However, the high-fidelity optical and physical simulations increase the hardware load and computational cost so RTX-based GPU is required. Another option is the more lightweight \textit{NVIDIA Isaac Lab}\cite{isaaclab}. Compared to Isaac Sim, it concentrates on the training of RL policies and supports large-scale, high-throughput multi-GPU distributed computations.

\renewcommand{\arraystretch}{1.25}
\begin{table*}[tbp]
    \caption{A Summary of the Advanced Autonomous Driving and MARL Simulators}
    \begin{center}
    \resizebox{\textwidth}{!}{
        \begin{tabular}{c|l|c|c|c|c|c}
\toprule
         & \multirow{2}{*}{\bf{Simulators}} & \multirow{2}{*}{\bf{Year}} &\bf{Last} & \multirow{2}{*}{\bf{State Description}} & \bf{MARL} & \multirow{2}{*}{\bf{Link}}\\
        & & & \bf{Update} & & \bf{Support} & \\
\midrule
        \multirow{19}{*}{\rotatebox{90}{\makecell{TrafficFlow-Oriented}}}
        & SUMO\cite{SUMO} & 2001 & \textcolor{blue}{2024-07} 
        & High-level states & \textcolor{green}{\CheckmarkBold} &
        \url{https://eclipse.dev/sumo/}\\
\cline{2-7}
        & Flow\cite{Flowsimulator} & 2018 & 2019-08 & High-level states & \textcolor{green}{\CheckmarkBold} & \url{https://flow-project.github.io}\\
\cline{2-7}
        & \multirow{2}{*}{Highway-env\cite{highwayenv}} & \multirow{2}{*}{2018} & \multirow{2}{*}{\textcolor{blue}{2024-04}} 
        & Binary image, grid map, & \multirow{2}{*}{\textcolor{green}{\CheckmarkBold}} &
        \multirow{2}{*}{\url{https://github.com/Farama-Foundation/HighwayEnv}}\\
        & & & & high-level states, etc. & & \\
\cline{2-7}
        & CityFlow\cite{zhang2019cityflow} & 2019 & 2020-12 
        & High-level states & \textcolor{green}{\CheckmarkBold} &
        \url{https://cityflow-project.github.io/}\\
\cline{2-7}
        & BARK\cite{BARK-semi} & 2020 & 2022-06 
        & Roadmap, high-level states, etc & \textcolor{red}{\XSolidBrush} &
        \url{https://github.com/bark-simulator/bark}\\
\cline{2-7}
        & MADRaS\cite{santara2021madras} & 2020 & 2020-10 
        & Image stream, high-level states & \textcolor{green}{\CheckmarkBold} &
        \url{https://github.com/madras-simulator/MADRaS}\\
\cline{2-7}
        & \multirow{2}{*}{SMARTS\cite{zhou2020smarts}} & \multirow{2}{*}{2020} & \multirow{2}{*}{\textcolor{blue}{2024-07}}  
        & BEV image, grid map, LiDAR, & \multirow{2}{*}{\textcolor{green}{\CheckmarkBold}} &
        \multirow{2}{*}{\url{https://github.com/huawei-noah/SMARTS}}\\
        & & & & high-level states, etc. & & \\
\cline{2-7}
        & \multirow{2}{*}{MetaDrive\cite{metadrive}} & \multirow{2}{*}{2021} & \multirow{2}{*}{\textcolor{blue}{2024-06}} 
        & RGB camera, high-level & \multirow{2}{*}{\textcolor{green}{\CheckmarkBold}} &
        \multirow{2}{*}{\url{https://github.com/metadriverse/metadrive}}\\
        & & & & states, etc. & & \\
\cline{2-7}
        & TBSim\cite{tbsim} & 2021 & 2023-10 
        & BEV image, etc. & \textcolor{red}{\XSolidBrush} &
        \url{https://github.com/NVlabs/traffic-behavior-simulation}\\
\cline{2-7}
        & TorchDriveSim\cite{torchdrivesim} & 2021 & \textcolor{blue}{2024-07} 
        & BEV image, etc. & \textcolor{red}{\XSolidBrush} &
        \url{https://github.com/inverted-ai/torchdrivesim}\\
\cline{2-7}
        & Intersim\cite{intersim} & 2022 & 2023-06 
        & Rasterized image, vectorized states & \textcolor{red}{\XSolidBrush} &
        \url{https://github.com/Tsinghua-MARS-Lab/InterSim}\\
\cline{2-7}
        & Nocturne\cite{nocturne} & 2022 & 2022-10 
        & High-level states & \textcolor{green}{\CheckmarkBold} &
        \url{https://github.com/facebookresearch/nocturne}\\
\cline{2-7}
        & \multirow{2}{*}{ScenarioNet\cite{scenarioNet}} & \multirow{2}{*}{2024} & \multirow{2}{*}{\textcolor{blue}{2024-05}} 
        & RGB camera, high-level states, etc. & \multirow{2}{*}{\textcolor{green}{\CheckmarkBold}} &
        \multirow{2}{*}{\url{https://github.com/metadriverse/scenarionet}}\\
        & & & & (depends on external datasets) & & \\
\cline{2-7}
        & \multirow{2}{*}{Waymax\cite{waymax}} & \multirow{2}{*}{2024} & \multirow{2}{*}{\textcolor{blue}{2024-03}} 
        & Roadmap, Bounding boxes, & \multirow{2}{*}{\textcolor{green}{\CheckmarkBold}} &
        \multirow{2}{*}{\url{https://github.com/waymo-research/waymax}}\\
        & & & & high-level states, etc. & & \\

\midrule

         \multirow{7}{*}{\rotatebox{90}{\makecell{Fidelity-Oriented}}}
        & TORCS\cite{torcs} & 2000 & 2020-02 
        & Image Stream & \textcolor{red}{\XSolidBrush} &
        \url{https://sourceforge.net/projects/torcs/}\\
\cline{2-7}
        & Gym-TORCS\cite{gymtorcs} & 2017 & 2017-02 
        & Image stream, high-level states & \textcolor{red}{\XSolidBrush} &
        \url{https://github.com/ugo-nama-kun/gym_torcs}\\
\cline{2-7}
        & \multirow{2}{*}{CARLA\cite{carla}} & \multirow{2}{*}{2017} & \multirow{2}{*}{\textcolor{blue}{2024-07}} 
        & RGB-D \& event camera, LiDAR, & \multirow{2}{*}{\textcolor{green}{\CheckmarkBold}} &
        \multirow{2}{*}{\url{https://github.com/carla-simulator/carla}}\\
        & & & & high-level states, etc. & & \\
\cline{2-7}
        & MACAD\cite{macad2020} & 2020 & 2023-01 
        & RGB camera & \textcolor{green}{\CheckmarkBold} &
        \url{https://github.com/praveen-palanisamy/macad-gym}\\
\cline{2-7}
        & \multirow{2}{*}{ISAAC Sim\cite{isaacgym2021}} & \multirow{2}{*}{2020} & \multirow{2}{*}{\textcolor{blue}{2024-06}} 
        & RGB-D camera, LiDAR, & \multirow{2}{*}{\textcolor{green}{\CheckmarkBold}} &
        \multirow{2}{*}{\url{https://developer.nvidia.com/isaac/sim}}\\
        & & & & high-level states, etc. & & \\
\cline{2-7}
        & \multirow{2}{*}{Vista\cite{vista1, vista2}} & \multirow{2}{*}{2020} & \multirow{2}{*}{2022-05}  
        & RGB-D \& event camera, LiDAR, & \multirow{2}{*}{\textcolor{green}{\CheckmarkBold}} &
        \multirow{2}{*}{\url{https://github.com/vista-simulator/vista}}\\
        & & & & etc. (depends on external datasets) & & \\
\bottomrule
        \end{tabular}
    }
    \label{table::simulators}
    \end{center}
\end{table*}

In table~\ref{table::simulators}, we summarize the key features of these representative simulators. Overall, we categorize simulators into two major types: traffic flow-oriented and fidelity-oriented. Traffic flow-oriented simulators prioritize parallelization and computational efficiency, often at the expense of accurate dynamics modeling and rendering precision, making them suitable for large-scale vehicle simulations. In contrast, fidelity-oriented simulators offer better optical rendering and dynamic accuracy with a smaller sim-to-real gap, but they consume significantly more computational resources. Specifically, we give out their website and code sources and list the supported sensors and information and note that most simulators provide abstracted observations, such as vectorized lane direction or the collision distance to the vehicle ahead. Here, we collectively refer to these processed features and representations along with the vehicle's kinematic information (like speed, acceleration or orientation) as the \textit{high-level states}. We also highlight the simulators that are still being maintained, and their updates are worth following. Researchers and practitioners should choose among these platforms based on their specific project needs, the complexity of the traffic scenarios, and the desired level of realism and interactivity in the simulations. We also notice that many unmentioned studies are showing fancy single-agent results on game engines\cite{GTAV, DeepDrive, airsim} and we acknowledge their contributions, but they can hardly adapt to MARL paradigm so they will not be covered in our survey.

\subsection{Datasets}
\label{subsec::dataset}
After decades of development, to effectively solve real-world driving problems, developers have collected tons of on-road data and established rich repositories with different sensors in various scenarios, such as KITTI\cite{kittidataset}, nuScenes\cite{nuscenes}, and Waymo Dataset\cite{waymodataset}. However, for decision-making problems, these datasets do not record real-time actions like accelerating, braking, steering, or actuator outputs, so they are typically used only for visual tasks or multi-modal sensor fusion. For example, although the IMU and GPS history is given out in BDD-100K\cite{yu2020bdd100k}, the information is on a different domain from human driver actions, and further policy adaption and transferring would be difficult. Moreover, most datasets address the autonomy of a single vehicle rather than collaboration between multiple AVs. Collecting  and aligning multi-vehicle data simultaneously is often expensive and difficult, so nowadays MARL paradigms are mostly verified in simulators. However, real vehicle decision data is still invaluable especially for imitation learning\cite{imitationl} and offline RL\cite{offlinerl}. Compared to data obtained in simulators, although theoretically we cannot interact for infinite time like simulation, the data distribution is closer to real driving.

To provide diverse interactive data for sequential decision-making, the INTERACTION dataset\cite{zhan2019interaction} investigates driving habits across different cultural backgrounds and collects bird's-eye-view data via hovering drones with fine-grained annotations. Based on the nations and road structures, it compiles 11 sub-datasets including highways, roundabouts, intersections, merges, and unstructured roads. Additionally, the dataset includes rare collision data and aggressive driving behaviors. Afterwards, the latest research proposes a new benchmark AD4RL\cite{lee2024ad4rl} based on Next-Generation Simulation~(NGSim) US-101 dataset\cite{alexiadis2004ngsim}. This benchmark provides 19 datasets from real-world human drivers after fine correction, value normalization, and alignment with partially observable Markov decision process. In other words, it can directly work as policy trajectories in the RL scheme.
Additionally, it offers 7 popular offline RL algorithms applied in 3 realistic driving scenarios. A unified decision-making process model is also attached for verification across various scenarios on Flow simulator\cite{Flowsimulator}, which serves as a reference framework for algorithm design. Recently, the development and supplement of datasets for the offline MARL approach in large-scale autonomous driving and traffic systems remain a work in progress. We consider this an up-and-coming area and will discuss it later (see Section~\ref{subsec::offline_data}).

\vspace{-5pt}
\subsection{Competitions}
\label{subsec::competition}
We notice that many recent competitions have been hosted during the international conference sessions, and here we address and appreciate the contribution of their organizers. These competitions provide a platform for researchers to showcase and compare their algorithms, drive the emergence of new techniques, and promote the application of MARL in real-world scenarios. 

The earliest multiple vehicles involved in competition can be traced back to the DARPA's Urban Challenge\cite{buehler2009darpa}. Although this challenge involved controlling only one vehicle, it required real-time controller adjustment based on a dynamic environment. In the past five years, an increasing number of companies and institutions have recognized MARL's potential and organized academic competitions based on relevant simulators and datasets. At DAI 2020, a multi-vehicle control competition was organized using the SMARTS\cite{zhou2020smarts} simulator. Participants were required to develop a parameter-sharing multi-agent model to control a group of agents to accomplish short missions in ramp, double merge, T-junction, crossroads, and roundabout. Instead of testing in the simulation, the DuckieTown\cite{paull2017duckietown} AI Driving Olympics~(AI-DO) competition on NeurIPS 2021 required participants to deploy real vehicles on scaled-down tracks. The vehicles had to navigate complex urban roads, avoiding pedestrians and other vehicles. The AI-Do competition marks a milestone in the practical deployment of embodied intelligence in complex MAS.
At NeurIPS 2022, Huawei's Noah's Ark Lab again organized competitions based on the SMARTS simulator, featuring online and offline reinforcement learning tracks. OpenDriveLab consecutively hosted multi-track autonomous driving competitions on CVPR 2023 and 2024. In the latest Autonomous Grand Challenge, the organizers provided the offline end-to-end autonomous driving scale competition and the online CARLA competition. For the former one, participants were required to develop motion planning algorithms for complex scenarios using offline data from Motional nuPlan\cite{caesar2021nuplan} dataset. The second track required the design of flexible policy learning methods in the CARLA\cite{carla} simulator. These competitions establish standardized benchmarks for evaluating different algorithms and offer a unified framework to compare performance and identify the most effective solutions. 

\section{Reinforcement Learning Preliminaries}
\label{prelim}

In this section, we will introduce the fundamental definitions and presentations of RL and MARL. We want to make sure the readers can comprehend the rest of this paper under a unified mathematical language.

\subsection{Deep Reinforcement Learning}
Reinforcement Learning is a classical approach to sequential decision-making problems. We typically use the Markov decision process~(MDP) to describe the decision-making procedure in RL paradigm, where the state distribution at the next time-step is only determined by the action and state of the current time-step and irrelevant to its history. 
Specifically, MDP can be denoted as a tuple $(\mathcal{S}, \mathcal{A}, \mathcal{R}, \mathcal{T}, \gamma)$, where $\mathcal{S}$ and $\mathcal{A}$ present the state space, action space and the stochastic observation space. 
$\mathcal{T}:\mathcal{S}\times\mathcal{A}\times\mathcal{S}\to\left[0, 1\right]$ is the transition probability under a given state-action pair and includes the uncertainty of the system.
$\mathcal{R}:\mathcal{S}\times\mathcal{A}\times\mathcal{S}\rightarrow r$ indicates the reward function which issues immediate reward value $r$ at each time-step and $\gamma$ presents the discount factor in the optimization objective. For a behavior trajectory, Equation~\eqref{obj} accumulates the discounted reward value $r_i$ at each time-step and our objective is to find an optimal policy to maximize the total return $\mathcal{G}_{t}$.
\begin{equation}
    \mathcal{G}_{t} =  r_{t+1} + \gamma r_{t+2} + \cdots =\sum\nolimits_{k=0}\gamma^{k}r_{t+k+1}
    \label{obj}
\end{equation}
\vspace{-15pt}
\begin{align}
    \mathcal{V}(s_t) &= 
    \mathds{E}_{s\sim\mathcal{S}}\left[\sum_{k=0}\nolimits\gamma^{k}r_{t+k+1} \right]\\
    & = \mathds{E} [ r_t + \gamma V(s_{t+1}) | s_t = s ]
    \label{v_function}
\end{align}
\begin{align}
    \!\!\!\mathcal{Q}(s_t, a_t) & \!=\!\mathds{E}_{s\sim\mathcal{S}, a\sim\pi_{\theta}}
    \left[ r_t + \gamma Q(s_{t+1}, a_{t\!+\!1}) | s_t\! = \!s, \!a_t \!= \!a\right]
    \label{q_function}
\end{align}
However, due to the stochastic environmental dynamics, we usually cannot quantify the quality of a state or action directly. For this issue, a common solution is estimating the expected subsequent reward given a specific state using the Bellman equation, referred to as the state value. The Eq.~\eqref{v_function} defines $\mathcal{V}$-function and evaluates the expected accumulative reward (i.e., return) of all subsequent states, which enables us to estimate state values through the dataset (or interactive trajectories) collected from simulation and real-world.
Another form of the Bellman Equation uses state-action pair $(s_t, a_t)$ and estimates the state-action value with $\mathcal{Q}$-function as Eq.~\eqref{q_function}. Here, the objective we evaluate is the expected return with a given state and action $a_t\sim\pi_{\theta}(s_t)$ sampled from policy $\pi$ with learnable parameters $\theta$. Essentially, the policy is a distributional function of actions with a given state. In deep RL, the $\mathcal{V}$-function, $\mathcal{Q}$-function, and policy $\pi$ are represented by neural networks. In general, the Bellman equation presents the accumulative reward given a behavior trajectory, and establishes the connection with the ground truth and the value estimation, so that we can the error between them to learn a better state evaluation via deep learning.

Meanwhile, to learn the policy safely while avoiding robot damage, we collect data (i.e., behavior trajectories) in the simulator and identify the policy that maximizes the accumulative reward. Based on how the policy learns from the collected data, the mainstream deep RL methods can be categorized into \textit{on-policy} and \textit{off-policy} RL. On-policy RL learns the value of the policy being carried out by the agent, meaning the policy used to make decisions is the same as the policy being improved. The advantage of on-policy RL is that it directly updates the target policy with the latest data and is more robust under sparse reward function\cite{drlsurvey2}. However, it requires more exploration during training as it must generate diverse experiences with the current policy. Contrarily, off-policy RL learns from different policies, which could be the old policy\cite{ppo} or a separate exploratory policy. Hence, it's more sample efficient as it can use data from multiple policies\cite{drlsurvey2}.

Another taxonomy is from their prerequisite and assumptions. Ideally, when the information of the environment during MDP can be fully known, Dynamic Programming\cite{howard1960dynamic} can solve the Bellman Equations via policy iteration or value iteration\cite{sutton2018reinforcement}. If the environment dynamics are partially known or unknown, the Monte Carlo method\cite{sutton2018reinforcement} estimates the unbiased value by implementing possible trajectories from the current state and calculating the return as Equation~\eqref{mc}. 
\begin{equation}
    \mathcal{V}(s_t) \leftarrow \mathcal{V}(s_t) + \alpha\left[\mathcal{G}_t-\mathcal{V}(s_t)\right]
    \label{mc}
\end{equation}
\begin{equation}
    \mathcal{V}(s_t) \leftarrow \mathcal{V}(s_t) + \alpha\left[\mathcal{G}_t + \lambda\mathcal{V}(s_{t+1})-\mathcal{V}(s_t)\right]
    \label{td0}
\end{equation}
Similarly, bootstrapping provides another estimation-based solution Temporal Difference~(TD) method. It predicts the future state from current time-step and as Equation~\eqref{td0}, we show the simplest method TD(0), where $\mathcal{G}_t + \lambda\mathcal{V}(s_{t+1})-\mathcal{V}(s_t)$ is TD-error at time $t$. In this way, TD-error facilities the update of the objective function in either on-policy form like SARSA\cite{rummery1994sarsa} or off-policy form like Q-learning\cite{watkins1992qlearning}.
Different from the value function-based methods above, another class of methods focuses on directly optimizing the parameters of the policy network, known as policy-based learning. In RL, the policy $\pi_{\theta}(a|s^{*}) = P(a|s^{*}, \theta)$ is essentially composed of a set of parameters $\theta$, which determines the probability of selecting an action under a given state $s^{*}$. 
\begin{equation}
    \mathcal{J}(\theta) = \mathbb{E}_{\tau\sim\pi_{\theta}} \left[ \sum\nolimits_{\Psi_t\in\tau} \Psi_t \right]
    \label{opt_obj}
\end{equation}
\begin{equation}
    \nabla_{\theta}\mathcal{J}(\theta) = \mathbb{E}_{\tau\sim\pi_{\theta}} \left[ \sum\nolimits_{(s_t, a_t, \Psi_t)\in\tau} \text{log}\left(\pi_{\theta}(a_t|s_t)\right) \Psi_t
    \right]
    \label{opt_grad}
\end{equation}
Accordingly, we aim to gain the optimal $\theta$, which maximizes our optimization objective in Equation~\eqref{opt_obj}, where $\Psi_t$ is a designable reward-related variable. In the simplest design, it is the reward itself. The roll-out trajectories sampled by policy $\pi_{\theta}$ are denoted as $\tau$. Under the mild conditions\cite{sutton1999policygradient}, we can obtain the gradient w.r.t. parameter $\theta$ via the key observation\cite{ilyas2020closer} as Equation~\eqref{opt_grad}. The REINFORCE\cite{williams1992reinforce} algorithm utilizes $\Psi_t = Q(s_t, a_t) - b$ with a learned coefficient $b$ to normalize Q-values. Similarly, the PPO algorithm\cite{ppo} proposes introducing the advantage function to modify and optimize the local approximation of the real return. Compared to value-based methods, policy gradient approaches circumvent curse of dimensionality arising from discretization in continuous action spaces. Essentially, policy gradient methods generate a probability distribution over actions applicable to discrete and continuous action spaces. 

The actor-critic architecture learns the policy and value function simultaneously. The actor (policy) makes decisions based on the current policy, and the critic (value function) evaluates the action selected by the actor. Actor-critic methods strike a balance between the high-bias value-based methods and high-variance policy-based methods and can efficiently operate in high-dimensional environments\cite{sac}. Due to space constraints, this paper will not undertake a deep dive into these categories and the implementation details. However, they are available in the reference if needed.

\subsection{Multi-Agent Deep Reinforcement Learning}
\label{subsec::madrl}
According to the assumption and agent settings, we can introduce diverse probabilistic models to represent the tasks. As an extension of MDP, Markov Game~(MG) presents the interaction process of a multi-agent system. MG is denoted a tuple $\left(\mathcal{N}, \mathcal{S}, \left\{\mathcal{A}^i\right\}, \left\{\mathcal{R}^i\right\}, \mathcal{P}, \gamma\right)$, where $\mathcal{N}=\left\{1, 2, \cdots ,N \right\}$ denotes the set of interacting agents and $\mathcal{S}$ is the global state from all agents with $i\in\mathcal{N}$. Similarly, $\left\{\mathcal{A}^i\right\}$ and $\left\{\mathcal{R}^i\right\}$ are the set of individual action and reward. Note that $\mathcal{A}^i$ is the action space of the $i$-th agent so the joint action space is $\mathcal{A}:=\mathcal{A}^1\times\mathcal{A}^2\times \cdots \times\mathcal{A}^N$. Like single-agent RL, $\mathcal{P}$ and $\gamma$ indicate the transition probability and discount factor. At time step $t$, each agent executes action $a^i_t\sim\pi^i$ according to the system state $s_t$, and then the system transits to the next state $s_{t+1}$ and obtain the reward $r^i_{t+1}:=\mathcal{R}^i(s_{t+1}|s_t, a^1_t, \cdots, a^N_t)$. Hence, we can obtain the value function for each agent like single-agent RL as Equation~\eqref{marl_v}.
\begin{equation}
    \mathcal{V}^{i}_{\pi^i, \pi^{-i}}(s_t) = 
    \mathbb{E}_{a^i\sim\pi^i}
    \left[
        \sum\nolimits_{k=0}\gamma^{k}r^i_{t+k+1}
    \right]
    \label{marl_v}
\end{equation}
\begin{equation}
     \mathcal{V}^{i}_{\pi^i_*, \pi^{-i}_*}(s_t) \ge \mathcal{V}^{i}_{\pi^i, \pi^{-i}_*}(s_t), \forall i \in \mathcal{N}
     \label{ne}
\end{equation}
Ideally, we hope that the policy of each agent $\pi^i$ should give out the best response to the joint policy of other agents $\pi^{-i}$ in the MAS. The Nash equilibrium~(NE)\cite{nash1950equilibrium} provides a more specific description for this case. Mathematically, we denote the joint policy in NE case as $\pi^i_* = \left\{\pi^1_*, \pi^2_*\cdots, \pi^N_*\right\}$ and then it can be described via value inequality~\eqref{ne}. Intuitively, the Nash Equilibrium indicates a situation with a joint policy, where no agent can change its policy to improve the performance unilaterally. However, the solutions representing NE in the state space are not unique in most cases\cite{gronauer2022marl2}. For a given joint policy $\hat{\pi}$ and any other policy $\pi$, if it satisfies $\mathcal{V}^{i}_{\hat{\pi}}(s) \ge \mathcal{V}^{i}_{\pi}(s)$ for all states $s\in\mathcal{S}$, and there is at least one state $s^*$ making  $\mathcal{V}^{i}_{\hat{\pi}}(s^*) > \mathcal{V}^{i}_{\pi}(s^*)$, the policy $\hat{\pi}$ is Pareto-optimal and it Pareto-dominates $\pi$. Accordingly, only a Nash Equilibrium without any other policy with greater value can be regarded as Pareto-optimal. 

In autonomous driving, vehicles' observations are limited by the field of view~(FoV) of the sensor and geographic location, so that they can only obtain incomplete information of environment. To this end, we typically modify the model by introducing the set $\mathcal{O}_i$ to denote observation space. Hence, We can define it as a decentralized partially observable MDP~(Dec-POMDP) with representation in a seven-element tuple $\left(\mathcal{N}, \mathcal{S}, \left\{\mathcal{A}^i\right\}, \left\{\mathcal{O}^i\right\}, \left\{\mathcal{R}^i\right\}, \mathcal{P}, \gamma\right)$. In MARL, the intricacies of inter-agent relationships influence the design of reward functions and training schemes. These relationships are typically classified into competitive, cooperative, and mixed. In competitive settings, agents strive to maximize their objectives, often at the expense of other agents. In other words, reward functions are typically zero-sum, where one agent's gain is equivalent to another's loss. Contrarily, cooperative agents are required to work collectively towards a common goal. In this case, the rewards are shared and distributed among agents based on their contribution to achieving the collective objective. Mixed interactions involve elements of both cooperation and competition; therefore, rewards in mixed scenarios are more complex and require a trade-off between individual and group incentives.

\begin{figure}[tbp]
    \centerline{
    \includegraphics[width=8.6cm]{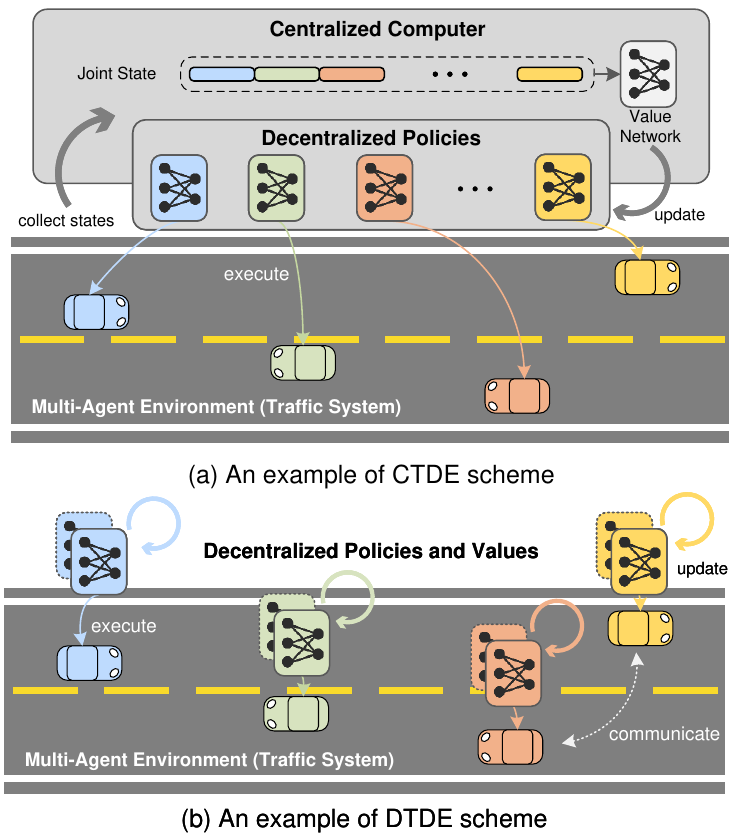}
    }
    \caption{The Centralized Training-Decentralized Execution~(CTDE) scheme versus Decentralized Training and Execution~(DTDE) scheme. In CTDE paradigm, it utilizes a central controller to concatenate all observations and distribute policies for all agents. In the DTDE paradigm, the agent computes the policy itself. According to hardware settings, they can communicate with each other and exchange information, or detect others' states by sensors.}
    \label{scheme}
\end{figure}
\subsection{Learning Schemes}
\label{subsec::scheme}
With the increasing number of agents, the complexities of state and action spaces would grow up exponentially, which causes the policy learning of MAS present a computational challenge. Generally, the entire process can be divided into two stages: Training and Testing. Training indicates the process where agents acquire data from interaction to obtain experience and update policies. After that, we evaluate the policy performance in the environment without any policy optimization, which is referred to as testing. Based on the classic categorization\cite{steels1995biology}, the training process can be broadly classified into two paradigms: centralized and decentralized. In centralized training, agents update their policies with communication and information exchange, while additional information is removed during testing. In contrast, decentralized training is conducted in a distributed manner, where each agent independently performs and develops individual policies without extra information and communication among agents. The centralized execution asks all agents to follow the command from one joint policy with an unconstrained and instantaneous information exchange\cite{gronauer2022marl2}. However, in real MAS, this strong assumption can hardly be achieved, and the agents are distributed various tasks and required to work independently. Therefore, we generally adopt decentralized execution, especially for autonomous driving cases. As shown in Figure~\ref{scheme}, mainstream MARL algorithms can be categorized into two paradigms: centralized training with decentralized execution (CTDE) and decentralized training with decentralized execution (DTDE). 

\subsubsection{CTDE Schemes}
Early centralized learning schemes are created to handle partial observable environments\cite{zhang2021marlsurvey1}. It assumes that a central controller (critic) collects the states and actions from all agents and updates the policies (actor) for them. Hence, the CTDE scheme with the Dec-POMDP model comes into being and has been applied in a deal of systems\cite{hansen2004decpomdp, rashid2020qmix, counterfactual}. In this case, we could use single-agent RL algorithms and toolkits to simplify the problem. However, for MAS systems with heterogeneous agents or multiple task objectives, this scheme can hardly learn feasible policies to meet various requirements. 

\subsubsection{DTDE Schemes}
In the fully decentralized paradigm, agents are trained and operate independently without access to global information. Therefore, it's more suitable to handle the decision-making and planning tasks for large-scale robotic systems\cite{copo, ruiqi2022pipo}. Classical DTDE approaches were limited by non-stationary and low efficiency, while recent research has significantly alleviated it via online evolution\cite{jaderberg2019human}, advanced value learning\cite{zhang2018fully, mao2022improving} and other techniques. Meanwhile, to deal with insufficient information from other agents, the decentralized MAS is generally established with a communication network where agents can exchange message or observe and estimate others' states. In the AD context, agents are typically configured to communicate with other vehicles within the signal range via Bluetooth or WiFi or to estimate their states through visual sensors.

\subsection{Issues of MARL}
\subsubsection{Non-Stationarity} 
Non-stationarity is a significant issue for decentralized MAS, where agents interact within a shared environment and update their policies synchronously. Consequently, each vehicle doesn't know the full environment dynamics, and the next state doesn't depend solely on its action and current state so it would break the Markov assumption\cite{gronauer2022marl2}. In the CTDE paradigm\cite{maddpg2017}, the centralized critic has access to all agents' observations and actions. Since only the actor computes the policy and the critic component can be removed during testing, agents in the CTDE scheme have fully decentralized execution. For independent policy learners, a naive approach is to let agents either ignore the presence of others or proceed under the assumption that the behaviors of others are static\cite{tan1993independent}. In this context, agents are independent learners, which enable the conventional single-agent RL algorithms for policy learning and have been proven to achieve excellent results on various benchmarks\cite{de2020independent}. However, in complex and stochastic environments, independent policy learning may result in sub-optimal performance or tend to exhibit a propensity for over-fitting to the policies of other agents, leading to a lack of generalizability in testing. To improve independent learning performance, researchers propose adopting different learning rates with shared rewards to achieve the optimal joint policy\cite{matignon2007hysteretic}. Another method involves the refinement of experience replay. Due to the non-stationarity of the environment, experience replay may store more irrelevant experiences to decentralized learning with the increasing time steps.  Importance sampling corrections for stable experience replay is also a solution, which adjusts the weights between the prior and the new experience under different environment dynamics. This approach has been proven to enhance the performance of independent learners in complex gaming and robotic environments\cite{foerster2017stabilising, papoudakis2019dealing, wu2021spatial}.

\subsubsection{Partial Observability}
In partially observable environments, agents do not have access to the full state of the environment. Instead, each agent receives only a local observation that provides incomplete or noisy information about the true state. To make effective decisions, agents must estimate or infer the underlying state of the environment from their partial observations. This often requires maintaining a a probability distribution over possible states, which adds computational complexity and uncertainty to the decision-making process. At the same time, in a partially observable setting, different agents might have different pieces of information about the environment. Coordinating actions effectively requires agents to share information or make decisions based on limited and potentially inconsistent knowledge.

For example, the policy learned with given sensors might require more work to find a workable latent representation and feasible policy from limited observations. To this end, a promising direction is using more complex architectures to enhance the representative capability of neural networks. Recent DRQN algorithm replaces the first layer of the DQN with a recurrent Long Short-Term Memory (LSTM) unit\cite{hochreiter1997lstm}. Combined with the concurrent experience replay trajectories mechanism, it mitigates the information limit on the hysteric Q-learning\cite{omidshafiei2017deep}. Inspired by this, researchers propose a weakly cooperative traffic model under traffic scenarios and apply an independent policy learning algorithm that utilizes a forgetting experience mechanism and a loose weight training mechanism to alleviate both partial observability and non-stationarity\cite{TVT2021independent}. Besides, MADDPG algotithm\cite{maddpg2017} with recurrent actor-critic has been demonstrated to learn low-variance stable policies in partially observable environments with various constraints of agent-level communication\cite{wang2020rmaddpg}.

\subsubsection{Credit Assignment} The credit assignment~(CA) problem is one of the crucial challenges in developing CTDE MARL. It involves determining how to allocate the global reward obtained through interaction with the environment\cite{credit_ai}. The most widely used method in such tasks is the shared team reward, which may lead to the problem of lazy agents, where only partial agents in the system work hard. Essentially, it's a tricky sub-optimal policy. To address this, Value Decomposition is the most widely-used solution, which divides the joint value function into individual agent value functions and then selects the best action for each agent. A straightforward linear decomposition represents the total reward function as a sum of each agent's reward function. Recent research focuses on solving credit assignments for heterogeneous agents through nonlinear assignments using a learnable weighted network\cite{credit_ai}. Another approach uses multi-agent policy gradient algorithms, which impose a monotonic constraint between the joint action value and individual policies\cite{CA_icml20, implicitCA_nips20, CA_nips21}. The recent SoTA method can find the global-optimal policy via polarized policy gradient\cite{CA_aaai23}, shortening the distance between multiple non-optimal joint action values. However, this problem still exists in large-scale and complicated heterogeneous robotic control problems. 

\subsubsection{Scalability}
The joint action space would increase exponentially with the increasing number of agents. Meanwhile, the number and density of agents are changing under specific real-world settings like the traffic flow of intersection, leading to variations in the dimensionality of observed information, which classic multi-layer perceptron~(MLP) or convolutional neural network~(CNN)-based\cite{lecun2015deep} value networks struggle to deal with. Although we can apply the solutions such as LSTM\cite{hochreiter1997lstm}, transformers\cite{vaswani2017attention}, or dimensionality reduction\cite{ruiqi2022pipo}, centralized methods require substantial computational resources, memory, and bandwidth to calculate and distribute their strategies after receiving action state information from each agent. A possible solution to the curse of dimensionality is independent learning. However, as mentioned previously, this approach must produce consistent results in non-stationary environments and may be limited by partial observability. In the context of autonomous driving, each agent is typically set up to exchange information with specific other nearby agents. For this, researchers propose a distributed Q-learning\cite{qdlearning}, assuming that each agent knows only its local actions and rewards, but agents can send their Q-values to their nearest peers and update their Q-networks locally. A scalable actor-critic method is established on exponential decay property with average reward within dynamic environments, which can handle the scaling state-action space size of local adjacent agents\cite{qu2020scalable}. In MAMBA\cite{egorov2022scalable}, researchers utilize model-based RL to sustain a world model\cite{ha2018worldmodel} for each agent during execution and generate efficient rollouts for training, removing the necessity of interaction with the environment.

\subsubsection{Communication Mechanism} 
As mentioned before, the communication mechanisms help MARL overcome issues of non-stationarity and partial observability. Besides, a better design of communication mechanisms contributes to reducing the lowest required hardware and bandwidth while enhancing learning efficiency. For the MAS in the real world, the agents are set to communicate with the adjacent or nearby agent. The communication relations can be predefined, changed, or even learned and updated during the process\cite{zhu2024survey, tro2024marlcomm}. Meanwhile, we can introduce the proxy to facilitate the collection, processing, and distribution of the message\cite{liu2020multi}. The message could contain both the past and current observations and actions, and it also can be compressed and processed using sequence networks or autoencoders\cite{kim2019message}. In some studies\cite{jiang2018learning, chu2020multi, kim2020communication}, the intended action or policy distributions can also be shared among agents and serve as additional input information for policy or value networks or both of them\cite{niu2021multi}. Note that some recent researches regard communication as a learnable process and focus on updating and adjusting communication protocols. Communication learning encompasses the communication policies and the content of messages\cite{zhu2024survey}. It can be implemented through backpropagating gradients from the communicatees\cite{sukhbaatar2016commback1, 2020commback2} or through an independent reinforcement learning thread\cite{foerster2016commrl1, gupta2023commrl2}.

\section{State-of-the-Art Methodologies}\label{method}
This section will introduce recently the most advanced MARL methodologies for motion planning and control of multi-vehicle systems. We cannot encompass all the related studies, but select representative techniques in this survey are sourced from reports published in the most influential conferences and journals. Furthermore, we encourage the researchers to report more relevant works to our website.

\subsection{Centralized Multi-Agent RL}
In the CTDE scheme, each vehicle has an independent policy network, and a core computer is set to merge and process the information from all vehicles. We first get the merged observation from all vehicles, evaluate the system state by a pre-defined global reward function, and then train the independent policies after credit assignment. PRIMAL\cite{2019primal} is a milestone work in centralized training for pathfinding. It assigns each agent an independent and fine-designed parameter-sharing actor-critic network and trains them with A3C\cite{2016a3c} algorithm. In this work, researchers illustrate that independent policies lead to selfish behaviors, and a hand-crafted reward function with a safety penalty is a good solution.
Additionally, there is a switch to allow agents to learn from interaction or expert demonstrations. The combination of reinforcement learning and imitation learning contributes to fast learning and alleviates the negative impact of selfish behaviors on the overall system. In this paper, a discrete grid world is defined, and the local state of each agent is set as the information of a 10$\times$10 block with the unit vector directed toward the goal. To verify the feasibility in the real world, the authors also implement PRIMAL on AVs in a factory mockup.

In MADDPG\cite{maddpg2017}, the authors propose the first generalizable CTDE algorithm based on deep deterministic policy gradient~(DDPG)\cite{lillicrap2016ddpg} with a toy multiple-particles environments. It provides an essential platform with easy vehicle dynamics to learn the continuous driving policies with continuous observation and action spaces under design-free scenarios and attracts many remarkable followers\cite{ruiqi2022pipo, qin2021learning}. Meanwhile, the combination of value function decomposition methods and CTDE scheme has achieved better scalability w.r.t. the number of agents and mitigates the impact of non-stationary on policy training, thereby improving performance in large-scale multi-agent systems\cite{sunehag2018vdn, rashid2020qmix}. These methods have been verified in complex scenarios like unsignalized intersections in Highway-Env\cite{IV2023_qmix, highwayenv}. Also, expert demonstration contributes to reducing the risk of converging to sub-optimal policies\cite{IV2023_qmix}. To verify the feasibility of deploying the CTDE approach in mapless navigation tasks, Global Dueling Q-learning~(GDQ)\cite{GDQ_turtlebot3} sets up an independent DDQN\cite{van2016ddqn} for each turtlebot3 in the MPE\cite{maddpg2017} to train policies and estimate values.
Additionally, they introduced a global value network that combines the outputs of the value networks of every agent to estimate the joint state value. This method has been proven to be more effective than normal value decomposition methods. Meanwhile, researchers also attempt to extend fundamental algorithms in single-agent RL such as PPO\cite{ppo} or SAC\cite{sac} to multi-agent tasks and provide many significant baselines like MAAC\cite{lowe2017maac} and MAPPO\cite{yu2022mappo}. In particular, MAPPO has been verified comprehensively on massive benchmarks and has systematic guidance of hyperparameter selection and training. To overcome the sim-to-real gap and deploy MAPPO on real robots, developers train a policy in the Duckietown-Gym simulator for following waypoints on the ground. The MAPPO policy network adopts recurrent neural network\cite{rnn} to recall the knowledge of the prior state and output the high-level target linear velocity and angular rate for each vehicle. Like most indoor navigation tasks, the optical track system captures the position and attitude of vehicles. With the linearized inverse kinetics, the executive low-level command of the vehicle can be obtained after domain adaptation. This work reveals how to deploy the CTDE scheme on real robots, and the engineering experience is valuable for future studies.

\subsection{Independent Policy Optimization}
\label{subsec::ipo}
Regarding practical deployment challenges such as communication, bandwidth, and system complexity, the fully decentralized system reduces communication overhead and bandwidth requirements by allowing agents to operate independently without constant coordination. Additionally, it is easier to deploy in environments with limited or unreliable communication infrastructure, lowers decision-making latency, and simplifies local computation for each agent. These factors make decentralized MARL a more practical and adaptable approach for real-world multi-agent applications. In recent years, Independent Policy Optimization~(IPO)\cite{2020ipg} has obtained increasing attention, and massive related approaches have been proposed. Concurrently, the complexity of the scenarios addressed in these studies and the scale of the agents involved have also been increasing synchronously, which reflects that decentralized learning matches the demands of large-scale autonomous driving in the real world more.

To solve the scalability issue in centralized schemes, MAPPER\cite{2020mapper} employs a decentralized actor-critic based on the A2C\cite{2016a3c} algorithm. Firstly, the local observations of the occupancy map are represented as a 3-channel image containing static scenes, dynamic obstacles, and planned trajectory information from A* planner\cite{1968astar}. These 3-channel observations are abstracted into a latent vector via a CNN, along with waypoint information abstracted by an MLP, input into shared fully connected layers. Later, two independent MLPs output action probabilities and value estimates, respectively. Besides, MAPPER employs an extra evolutionary algorithm to eliminate bad policies during the optimization process. Compared with PRIMAL\cite{2019primal}, MAPPER can learn faster and handle dynamic obstacles more effectively in large-scale scenarios. 
Another work scalability is G2RL\cite{2020G2RL}, a grid map navigation method that can be used for any arbitrary number of agents. Similarly, it leverages A* to provide each agent with a global guiding path. Meanwhile, the local occupancy map is input into a local DDQN\cite{van2016ddqn} planner to capture local observation and generate a corrective command to avoid dynamic obstacles. Since there is no need for communication between agents, this method does not require consideration of communication delays and can be extended to any scale.

As the successor to PRIMAL, PRIMAL$_2$\cite{2021primal2} retains the same hierarchical structure, i.e., an A* planner generating global paths and agent training guided by A3C and imitation learning. The key difference lies in PRIMAL$_2$'s fully decentralized training approach, which enhances its flexibility in handling structured and high-density complex scenarios. Like MAPPER, it adopts an 11$\times$11 observation range and splits observations into multi-channel image inputs. The first 4 channels include static obstacles, the agent's own goal point, other agents' positions, and other agents' goal points. Channels 5-8 provide the local path from A* and the positions of other agents at three future timesteps within the observation range. The final 3 channels offer the X and Y coordinate offsets of corridor exits and a boolean state indicating whether other agents are blocking the path. The finer observation inputs allow PRIMAL$_2$ to effectively address the agent deadlock issues in high-density, complex occupancy grids with shorter generated paths than its predecessor.

The aforementioned methods are developed for structured occupancy grids with discrete action spaces and applicable to automated ground vehicles in structured warehouses and freight terminals. While there are differences from real traffic systems, these methods remain inspirational for subsequent work. Other decentralized learning studies are conducted on more advanced continuous benchmarks\cite{maddpg2017, carla, metadrive}. For instance, in PIPO\cite{ruiqi2022pipo}, researchers develop an end-to-end motion planning scheme using the permutation-invariant property of MAS with a graph neural network. They defined a progressively larger continuous scenario in MPE with various static obstacles. During training, the random permutation of observed states of other agents enhanced the feature representation of the actor-critic network. We note that there are numerous excellent and representative decentralized training schemes, but we categorize them under other subtopics and will elaborate on them in the following sections.

\subsection{Learning with Social Preference}
\label{subsec::soc_pre}
\begin{figure}[tbp]
    \centering
    \includegraphics[width=8.6cm]{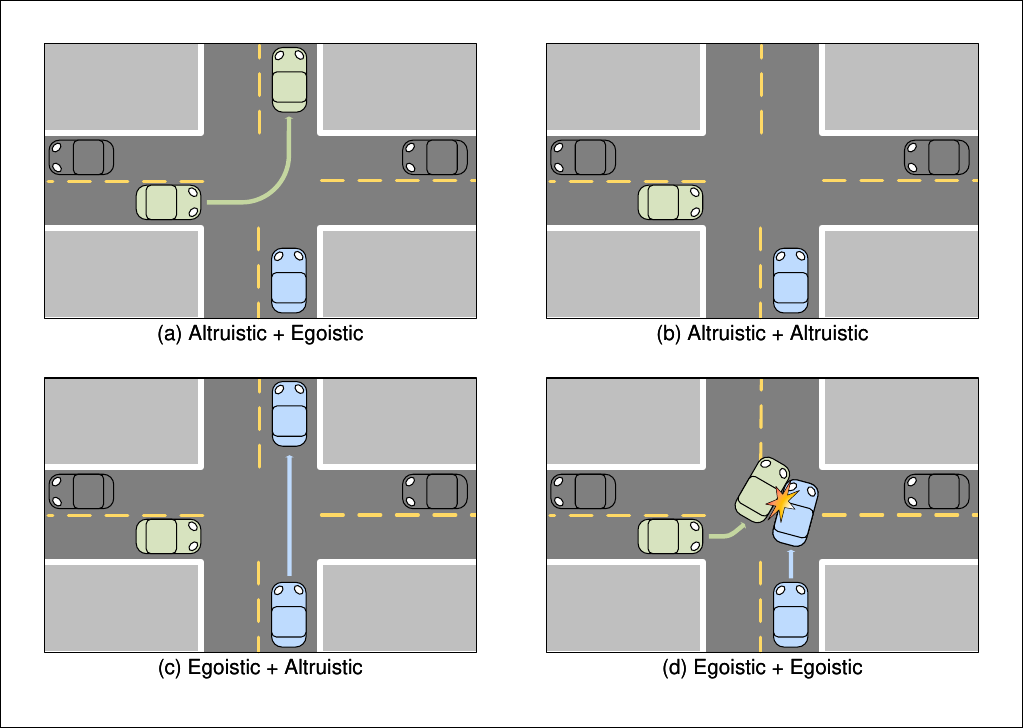}
    \caption{Four examples at free intersection to show how social preference effects the behaviors of AVs. In (a) and (c), the combination (i.e. one egoistic + one altruistic) is healthy; in (b), two altruistic cars would wait for each other; in (d), two egoistic cars would crash into each other.} 
    \label{fig:social}
\end{figure}
Although independent policy learning is feasible in many tasks, it would lead to each agent being self-centric\cite{copo} when the interests of multiple agents conflict, the pure egoistic independent policy learning may fail. Therefore, an important issue is balancing the agents' egoism and altruism. In Fig.~\ref{fig:social}, we give a toy example to illustrate how social preference affects the agents' behaviors. If the agents cannot balance their altruistic and egoistic behaviors, these two would crash or get stopped by each other. Hence, social behaviors and preferences should be considered in policy learning\cite{social_PNAS}. To find a mathematical presentation of social preference, in the early work, researchers first propose to use a trigonometric function like Eq.~\eqref{eq::copo_co} to balance the individual and global reward, which inspires afterward studies\cite{copo, sapo}. 

\begin{figure}[tbp]
    \centering
    \includegraphics[width=8.6cm]{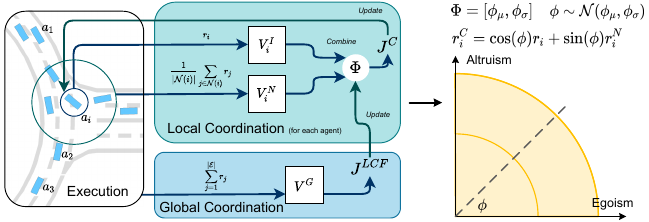}
    \caption{The architecture of Coordinate Policy Optimization\cite{copo}. The bi-level training process enables to balance egoism and altruism.} 
    \label{fig:copo}
\end{figure}

Afterward, as the representative work of driving with social preference, Coordinate Policy Optimization\cite{copo}~(CoPO) draws inspiration from the self-driven particle systems in nature like fish and bird flocks and implements a decentralized MARL method with heterogeneous vehicle settings in MetaDrive\cite{metadrive}. It proposes hierarchical coordination for policy optimization to trade off the egoism and altruism of the policy. It utilizes the IPO for each vehicle within the traffic system to avoid the credit assignment problem. More specifically, by introducing a local coordination factor~(LCF) in the training process, the agent seeks the optimal policy to maximize the averaged reward from all adjacent agents in its observable range. As Fig.~\ref{fig:copo}, for the i-th agent $a_i$ in the system with time-invariant observed adjacent agents $N(i, t)$ in the range $d_n$, the agent should balance its ego reward $r_i$ and the average reward $r^N_i(t)$ as shown in Eq.~\eqref{eq::copo_co}.
\begin{equation}
    r^N_i(t) = \frac{1}{|N(i, t)|}\sum\nolimits_{j\in N(i, t)}r_j(t)
    \label{eq::copo_rN}
\end{equation}
\begin{equation}
    r^C_i(t) = \cos(\phi)r_i(t) + \sin(\phi)r^N_i(t), \phi \in \left[-90^\circ, 90^\circ\right]
    \label{eq::copo_co}
\end{equation}
LCF $\phi$ indicates the specific altruism of the policy under a certain scenario, so the major difficulty of local coordination comes from selecting optimal LCF to maximize the global reward. Hence, in the global coordination, the author presents LCF as a Gaussian distribution 
$\phi\sim\mathcal{N}(\phi_{\mu}, \phi_{\sigma})$ and the global objective is to find out the optimal policy with maximum accumulative rewards 
$J^G(\theta_1, \theta_2, \cdots) = \mathbb{E}_\tau\left[ \sum\nolimits_{i\in N} \sum\nolimits_{t=0}^T r_i(t) \right] $. 
To enable LCF to be learnable, the authors transfer the global objective into individual objective following with an easy factorization technique\cite{rashid2020qmix} and then derive the gradient from Eq.~\eqref{eq::copo_gradient} to~\eqref{eq::copo_gradient2}, where $\theta^o_i$ and $\theta^n_i$ denote the old new updated policy network of agent $i$.
\begin{equation}
    \bigtriangledown _{\Phi} J^G_i(\theta_i^{n})=
    \bigtriangledown _{\theta_i^{n}} J^G_i (\theta_i^{n}) \cdot
    \bigtriangledown _{\Phi} \theta_i^{n}
    \label{eq::copo_gradient}
\end{equation}
\begin{equation}
     \bigtriangledown _{\theta_i^{n}} J^G_i (\theta_i^{n}) = 
    \mathbb{E} \left[  
        \bigtriangledown_{\theta_i^{n}} \min (\rho A^G, \text{clip}(\rho, 1-\epsilon, 1+\epsilon)A^G) 
        \right]
    \label{eq::copo_gradient1}
\end{equation}
\begin{equation}
    \bigtriangledown _{\Phi} \theta_i^{n} = 
    \alpha \cdot \mathbb{E} \left[
         \bigtriangledown_{\theta_i^{o}} \log \pi_{\theta^o_i(a_i|s)}
         \bigtriangledown_{\Phi}A_{\Phi, i}^C
    \right]
    \label{eq::copo_gradient2}
\end{equation}
Here, $\Phi=[\phi_{\mu}, \phi_{\sigma}]$ denotes the mean and variance of LCF and is learnable, and $\alpha$ denotes the learning rate. In Eq.~\eqref{eq::copo_gradient1} and \eqref{eq::copo_gradient2}, $A^G$ and $A^C_{\Phi, i}$ is the advantage of global and locally coordinated reward respectively following the idea in \cite{ppo}. Since Eq.~\eqref{eq::copo_gradient1} has no relevant term of $\Phi$, it can be regarded as a constant in the LCF objective. Hence, the objective of learning can be established as follows:
\begin{equation}
    J^{LCF}(\Phi) = \mathbb{E}  
    \left[\bigtriangledown _{\theta_i^{n}} J^G_i (\theta_i^{n})\right] \cdot
    \left[\bigtriangledown_{\theta_i^{o}} \log \pi_{\theta^o_i(a_i|s)} A_{\Phi, i}^C \right]
\end{equation}

Introducing a learnable factor to present social preference outperforms independent learning and mean-field policy optimization. It has also inspired many valuable discussions in this field about learning independent policy with sociological design and led to other excellent works. In subsequent research, people leverage more advanced language models like Transformer\cite{vaswani2017attention} to process social preference. For instance, in Social-Attention Policy Optimization~(SAPO)\cite{sapo}, the authors introduce a multi-head attention layer to select the most interactive agent of the ego vehicle. In this work, the state of the ego vehicle is embedded into queries and compared to all the keys, and the interactive agent could be selected from the non-zero value index. After Gumbel Softmax\cite{jang2016categorical} with gathered one-hot values, it obtains a compact permutation-invariant feature as the residual input of observation. Essentially, social coordination is a unique design of the reward function to balance collective and individual interests and align with the operational logic of real-world society.

\subsection{Safe and Trust-Worthy Learning}
\label{subsec::safety_rl}
Safety is integral and the first priority to the deployment of autonomous driving systems, as they directly impact the reliability and people's lives of AVs. Recent RL researchers put massive efforts into ensuring the learned policy would never cause safety issues in the exploration process and after deployment. Specifically, inspired by \cite{brunke2022safe_survey}, we categorize existing safety standards and methods in MARL into three types. First, soft safety guarantees involve designing safety penalty terms to reduce the probability of dangerous behavior. With fine-tuned rewards, the learning algorithm can be guided to prioritize safety alongside other performance metrics. However, although they have been proven to effectively improve safety performances in MAS, the limitation of soft guarantees is that they rely on the assumption that the reward function can accurately capture all safety aspects, which is often challenging in complex environments. The second is the probabilistic guarantees happening in the optimization process. For example, some recent MARL algorithms leverage the Lagrange constraints\cite{ruiqi2022pipo}  or safety threshold during policy optimization process\cite{gu2023safe, ying2024scalable}. Essentially, this improves policy gradient and helps avoid dangerous exploration behaviors. However, since the policy is still represented as a probability distribution, we cannot obtain a clear, explainable, and stable safety boundary for this method. Meanwhile, the vital safety constraints in real-world driving are instantaneous and deterministic\cite{zhao2023statewise}. For example, collision avoidance is a state-wise instantaneous constraint that only depends on the current state of the system rather than historical trajectories or random variables.

\begin{table*}[htbp]
        \caption{Recent Multi-Agent Reinforcement Learning Methodology  in Autonomous Driving \\ and Intelligent Transportation System (ranked by year)}
        \begin{center}
        \begin{tabular}{l|c|c|c|c|c|c|c}
        \toprule
        \multirow{2}{*}{\bf{Research}} & \multirow{2}{*}{\bf{Year}} & \bf{ Training} & \multirow{2}{*}{\bf{Simulator}} & \bf{Maximum} & 
        \bf{Vehicle} & \bf{Action Space}  & \bf{Real-World}\\
        & & \bf{Scheme} & & \bf{Agents} & \bf{Kinetics} & \bf{Description} & \bf{Experiment}\\
\midrule

        MAMPS\cite{zhang2019mamps} & 2019 & Centralized & MPE & 4
        & Omnidirection & Continuous, 2 dims &  \textcolor{red}{\XSolidBrush}\\
\hline

        PRIMAL\cite{2019primal} & 2019 & Centralized & Grid Map & 1024 
        &  Grid model & Discrete, 5 dims &   \textcolor{green}{\CheckmarkBold}\\
\hline

        \multirow{2}{*}{MAPPER\cite{2020mapper}} & \multirow{2}{*}{2020} & \multirow{2}{*}{Decentralized} & \multirow{2}{*}{Grid Map} & \multirow{2}{*}{150} 
        & \multirow{2}{*}{Grid model} & Discrete, 9 dims &  \multirow{2}{*}{\textcolor{red}{\XSolidBrush}}\\
        & & & & & & (add $\nearrow, \searrow, \swarrow, \nwarrow$ ) & \\
\hline

        G2RL\cite{2020G2RL} & 2020 & Decentralized & Grid Map & 128 
        & Grid model & Discrete, 5 dims &  \textcolor{red}{\XSolidBrush}\\
\hline

        \multirow{2}{*}{MACAD\cite{macad2020}} & \multirow{2}{*}{2020} & \multirow{2}{*}{Decentralized} & \multirow{2}{*}{MACAD} & \multirow{2}{*}{3}
        & \multirow{2}{*}{Bicycle model} & Discrete, 9 dims & \multirow{2}{*}{\textcolor{red}{\XSolidBrush}}\\
        & & & & & & (predefined action tuples)& \\
\hline        
        CoPO\cite{copo} & 2021 & Decentralized & MetaDrive & 40 
        & Full-vehicle model & Continuous, 2 dims & \textcolor{red}{\XSolidBrush}\\
\hline

        GDQ\cite{GDQ_turtlebot3} & 2021 & Centralized & MPE & 8 
       & Omnidirection & Continuous, 2 dims & \textcolor{green}{\CheckmarkBold}\\
\hline

        PRIMAL$_2$\cite{2021primal2} & 2021 & Decentralized & Grid Map & 2048 
        & Grid model & Discrete, 5 dims & \textcolor{green}{\CheckmarkBold}\\
\hline

        PIPO\cite{ruiqi2022pipo} & 2022 & Decentralized & MPE & 512 
       & Omnidirection & Continuous, 2 dims & \textcolor{red}{\XSolidBrush}\\
\hline

        \multirow{2}{*}{SSMAQL\cite{TITS_social}} & \multirow{2}{*}{2022} & \multirow{2}{*}{Decentralized} & \multirow{2}{*}{Gym} & \multirow{2}{*}{4}
        & \multirow{2}{*}{Bicycle model} & Discrete, 5 dims & \multirow{2}{*}{\textcolor{red}{\XSolidBrush}}\\
        & & & & & & (high-level action)& \\
\hline        

        \multirow{2}{*}{Duckie-MAAD\cite{mappo_duck}} & \multirow{2}{*}{2022} & \multirow{2}{*}{Centralized} & \multirow{2}{*}{Gym} & \multirow{2}{*}{3}
        & \multirow{2}{*}{Omnidirection} & Discrete, 4 dims & \multirow{2}{*}{\textcolor{green}{\CheckmarkBold}}\\
        & & & & & & (high-level action)& \\
\hline  
        \multirow{2}{*}{QMIXwD\cite{IV2023_qmix}} & \multirow{2}{*}{2023} & \multirow{2}{*}{Centralized} & \multirow{2}{*}{Highway-env} & \multirow{2}{*}{4}
        & \multirow{2}{*}{Bicycle model} & Discrete, 3 dims & \multirow{2}{*}{\textcolor{red}{\XSolidBrush}}\\
        & & & & & & (high-level action)& \\
\hline  

        TIRL\cite{2023TIRL} & 2023 & Decentralized & Grid Map & 64 
        & Grid model  & Discrete, 5 dims & \textcolor{green}{\CheckmarkBold}\\
\hline

        
        \multirow{2}{*}{SAPO\cite{sapo}} & \multirow{2}{*}{2023} & \multirow{2}{*}{Decentralized} & \multirow{2}{*}{SMARTS} & \multirow{2}{*}{4}
        & \multirow{2}{*}{Bicycle model} & Discrete, 3 dims & \multirow{2}{*}{\textcolor{red}{\XSolidBrush}}\\
        & & & & & & (high-level action)& \\
\hline  

        \multirow{2}{*}{CS-MADDPG\cite{zheng2024csmaddpg}} & \multirow{2}{*}{2024} & \multirow{2}{*}{Centralized} & \multirow{2}{*}{Highway-Env} & \multirow{2}{*}{2}
        & \multirow{2}{*}{Bicycle model} & Discrete, 5 dims & \multirow{2}{*}{\textcolor{red}{\XSolidBrush}}\\
        & & & & & & (high-level action)& \\
\hline  

        \multirow{2}{*}{CAVMARL\cite{MARL_CBF2024TITS}} & \multirow{2}{*}{2024} & \multirow{2}{*}{Decentralized} & \multirow{2}{*}{CARLA} & \multirow{2}{*}{30}
        & \multirow{2}{*}{Bicycle model} & Discrete, 3 dims & \multirow{2}{*}{\textcolor{red}{\XSolidBrush}}\\
        & & & & & & (high-level action)& \\
\hline  

        MFPG\cite{2024mfpg} & 2024 & Decentralized & ROS & 8 
        & Omnidirection & Continuous, 1 dim & \textcolor{green}{\CheckmarkBold}\\
\hline

        CPO-AD\cite{2024macpo_ad} & 2024 & Centralized & CARLA & 22 
        & Longitude only &  Continuous, 1 dim &  \textcolor{red}{\XSolidBrush}\\
        \bottomrule
        \end{tabular}
        \label{tab::method}
        \end{center}
\end{table*}

The third and safest approach is using hard safety boundaries to apply instantaneous strong corrections for agents' actions. For instance, researchers propose to learn the centralized shielding\cite{elsayed2021safe} from the joint action to correct any unsafe action in MAS at any risky time. Alternatively, combined with model predictive control, researchers propose a multi-agent model predictive shielding algorithm that provably guarantees safety for any policy learned from MARL\cite{zhang2019mamps}. However, due to the centralized setting, these methods cannot scale to the massive number of agents. Another widely-applied method for safety guarantee is control barrier function~(CBF)\cite{2014cbf, tro2021cbf}. Assuming known explicit dynamics, early CBF-based safe controllers could ensure the safety of control strategies in simple tasks and environments\cite{2014cbf}. However, real-world autonomous driving introduces complex nonlinear dynamics and intricate agent-level interactions, which lead the hand-crafted CBF to become infeasible\cite{2024macpo_ad}. Therefore, some studies have proposed incorporating additional neural networks to extract CBF for MAS and visualize it to interpret the safety boundaries\cite{qin2021learning, MARL_CBF2024TITS, tro2023cbf}. For instance, MDBC\cite{qin2021learning} introduces neural barrier certificates for each agent and achieves scalable and super safe decentralized controllers for up to 1024 AVs and drones in particle environments. In the latest CAVMARL\cite{MARL_CBF2024TITS}, researchers establish a safe action mapping via CBF-based quadratic programming. This controller leverages truncated Q-function to ensure the scalable joint state-action estimation under a centralized scheme and generates steering angle and acceleration with mathematically provable safety certificates.

\subsection{Methodological Summary}
As shown in Table~\ref{tab::method}, we collect representative works on MARL in outdoor autonomous driving, traffic system control, and structured scene transportation in the past five years. Meanwhile, we list their taxonomy, the maximum number of agents, the simulators, and whether real-world experiments are conducted. Here, we note that the action settings can be completely different even with the same simulation type. For example, in PRIMAL and PRIMAL$_2$, the agent's actions are set as ($\uparrow, \longrightarrow, \downarrow, \longleftarrow, *$), representing four movements in the horizontal and vertical directions in a 2D grid map, along with staying in place. In contrast, MAPPER adds four additional diagonal movements ($\nearrow, \searrow, \swarrow, \nwarrow$ ) for the agents. Additionally, we find that many studies adopt predefined high-level action commands to simplify tasks. The policy network outputs discrete values that map to corresponding preset actions, and then a low-level controller takes the actions, generates commands, and sends them to the actuators. Two other specific examples are MFPG\cite{2024mfpg} and CPO-AD\cite{2024macpo_ad}. They preset a low-level unidirectional control mapping and only consider the movement of AVs in one direction.

Furthermore, we summarize three trends from past studies in this field. First, early research is limited by algorithm diversity and simulator performance and focuses more on centralized MARL in grid maps. However, recent studies have discussed the potential of decentralized approaches with more complex continuous observation. Second, only a few studies conduct real-world experiments and use only discrete simulators and few agents. That is what the future works could improve. Third, the newer studies adopt more complicated designs and integrate more methods from other fields, such as data compression and machine vision.

\section{Open Questions and Challenges}
\label{challenge}
In this section, we present the main challenges in MARL. Note that the problems faced by the CTDE and DTDE schemes are different, and though some feasible solutions have been proposed to solve these issues, they are still not unique and perfect. We hope that readers can become aware of their existence and properties in advance, thereby gaining a better understanding of the motivation and technical innovation from subsequent advanced methodologies.

\subsection{Multi-modal Information}
Autonomous driving is a sequential decision-making process that leverages multi-modal information. Compared to MLPs in the original algorithm designs, recent research focuses more on designing more complicated neural network modules to learn better representations from multiple sensor information and to suit time series information.

\subsubsection{Multi-Sensor Fusion and Integration}
AVs acquire kinematic sensors like GPS and IMU, visual information from RGB-D cameras, LiDAR, and event cameras, and output executable commands to actuators. The multi-sensor information fusion enhances system safety redundancy, ensuring the vehicle can continue operating safely even if one sensor fails~\cite{singh2023transformer}. However, sensors with different principles possess various properties in real-world scenarios. For instance, like human drivers, RGB cameras capture the 3-channel information from reflected rays on the imaging plane and are cost-effective. However, RGB cameras are sensitive to light intensity and would overexpose in strong light or be covered by thermal noise in dark scenarios.
Meanwhile, they are unsuitable for capturing high-speed objects for motion blur. Conversely, event cameras utilize parallelized pixels to capture brightness changes caused by high-speed moving objects, but they contain no color information\cite{eventsurvey}. LiDAR or depth camera can provide precise 3D spatial information. Also, information on kinematics and dynamics is significant for motion planning. Hence, it's challenging to merge these diverse data varying in formats, physical concepts, dimensions, and magnitudes. Early sensor fusion happens before the sensor information is input into the neural network. It concatenates the multi-modal information directly and relies on neural networks to extract helpful features\cite{ruiqi2022residual, copo, zheng2024csmaddpg}. This method is easy to implement but may result in the loss of geometric information and lead to more complex feature extraction. Contrarily, in the middle fusion paradigm, independent encoders extract key features from various modalities and squeeze them into latent vectors\cite{2021primal2, ruiqi2022pipo, 2020mapper}. These feature vectors are concatenated along the feature dimension into a long vector as the network input. Current research focuses on transferring the attention mechanism of language models to visual tasks and capturing contextual relationships between different modalities. Recent advances in sequential modeling and audio-visual fusion demonstrate that Transformer\cite{vaswani2017attention} is competent in modeling the information interaction for sequential or cross-modal data\cite{bai2022transfusion, chen2023futr3d}.

\subsubsection{Learning Representations}
Massive approaches are involved in discussing the representation of the multi-modal information for vehicle sensor systems. Classic autonomous driving algorithms rely on HD maps, using diverse formulations and topological signs to represent map structures like segmented maps, vectorized centerlines, and landlines. For vehicle-centered maps, some studies directly input the sensor outputs like 3-channel pixel information or distances from LiDAR to network\cite{ruiqi2022residual, ruiqi2022pipo, 2016e2edriving}. 2D grids or occupancy maps are a classical representation of vehicle-centered maps, which have been used in navigation in complex structured environments\cite{2019primal, 2020mapper, 2021primal2} and high-speed racing\cite{2022latent}. Alternatively, with depth cameras and LiDAR, the 3D scene can be reconstructed through SLAM\cite{zhang2014loam} or Neural Radiance Field~(NeRF)\cite{mildenhall2020nerf} and is represented in a 3D occupancy grid. However, high-resolution grids can lead to unacceptable computational consumption\cite{ailab2023e2esurvey}. Recent research\cite{liang2022bevfusion, MIT2023bevfusion2, wang2024rcdn} proposes to unify multi-modal features in the shared bird's-eye view (BEV) representation space, which preserves both geometric and semantic information and suit for downstream tasks like navigation and obstacle avoidance. Although various representations offer potential designs for integration with MARL, determining which approach is the most effective remains an open and unresolved question. Additionally, balancing hardware costs with algorithmic performance and identifying which information is essential are crucial considerations for representation design.

\begin{figure}
    \centering
    \includegraphics[width=8.6cm]{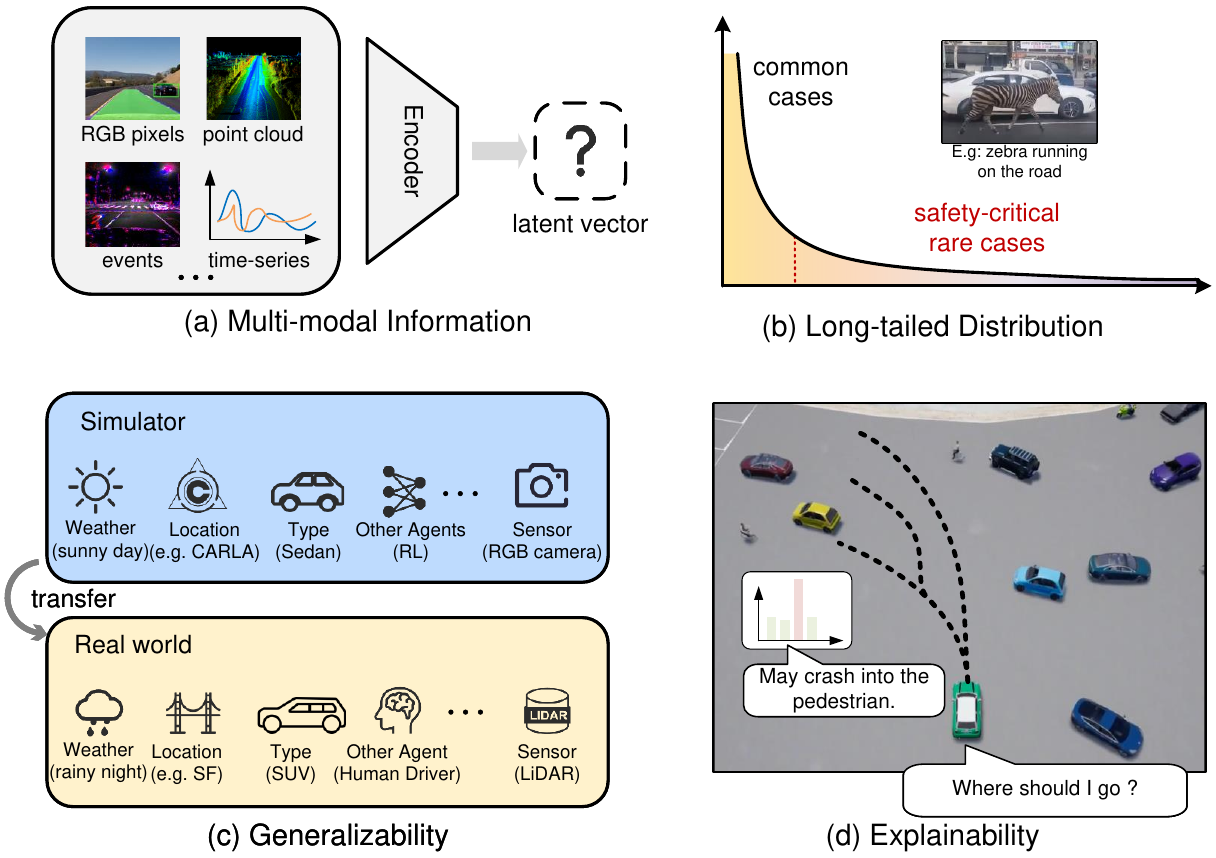}
    \caption{Some examples of open questions and challenges. (a) AVs have to process multi-modal information from various sensors. (b) The long-tail distribution makes covering all scenarios in simulators or datasets difficult. (c) The learned policy should be generalizable to different environments and agents, and overcome the sim-to-real gap. (d) The learned policy should give a clear explanation for its action.}
    \label{fig:challenge}
\end{figure}

\subsection{Robustness and Generalizability}
Robustness and generalizability are critical factors for the effectiveness of MARL algorithms in autonomous driving. Robustness allows the vehicle to safely navigate a wide range of real-world conditions, while generalizability ensures the system can be applied broadly across different environments and scenarios. These qualities are fundamental to advancing autonomous driving technologies and achieving practical deployment in diverse and unpredictable real-world settings.

\subsubsection{Long-tailed Problem}
\label{subsubsec::longtail}

Autonomous vehicles are supposed to handle a broad spectrum of driving conditions, which include unexpected road obstructions, erratic behaviors from other drivers, unusual weather, and sudden mechanical failures. Numerous unpredictable corner cases lead to a gap in the vehicle's ability to respond effectively. As shown in Fig.~\ref{fig:challenge}, long-tailed problem\cite{2023longtailsurvey} indicates a tricky data distribution of learning method, i.e., AVs' policies training can cover most common cases, but may fail to make safety-critical decisions in the rare scenarios that it has never seen before.

Hand-craft test bases\cite{carla, SUMO, 2020handcraftscene, suo2021trafficsim} and heuristic methods\cite{gambi2019autotest, 2019didwe} are leveraged to generate a broader range of traffic scenes. However, these generated scenarios demand significant human effort and domain expertise to create effective rules and often fail to capture the complexity of real-world traffic and structures accurately. Recently, researchers have been trying to achieve automatic traffic scene generation from real-world datasets to a unified representation and synthesize long realistic trajectories for RL\cite{lan2023trafficgen, scenarioNet}. By exposing autonomous driving systems to these simulated environments, developers can ensure that the vehicles learn to manage unusual situations. This method complements real-world testing and helps build a more robust and adaptable system.
Meanwhile, continual learning\cite{2024continuallearning} and transfer learning\cite{2020transferlearning} are practical tools to address long-tailed problems. These approaches enable MAS to learn from new experiences and update policies to incorporate novel situations so the vehicle can deal with everyday and rare scenarios throughout its operational life. In general, efficiently establishing realistic scenarios covering the long-tailed distribution of these rare and critical conditions is still a long-term challenge for autonomous driving.

\subsubsection{Sim-to-Real Gap}
Although it's useful for policy learning and pre-hoc testing, simulated environments often fail to capture real-world scenarios' full complexity and variability. Factors such as weather, road variations, and unpredictable behavior of other road components are difficult to replicate accurately in simulations\cite{2024sim2real}. This disparity can lead to autonomous driving systems performing well in simulated tests but failing to handle the nuances of real-world driving. First, simulators typically use idealized models of sensors and actuators, which do not account for the imperfections and limitations of real devices. This can lead to discrepancies in sensor readings and actuator responses, impacting the system's performance when deployed in real-world scenarios. Latency is another critical factor that differentiates simulations from real driving. In simulations, information transfer and processing times can be minimized or overlooked. However, in real-world applications, delays in communication, sensor data processing, decision-making, and actuation can significantly affect the system's responsiveness\cite{2020latency}. Although we can arbitrarily define the frequency and delay of sensors and actuators in the advanced simulators\cite{carla, isaacsim}, it is challenging to model the fluctuations and disturbances of signal in the real world.
Additionally, for reinforcement learning, predicting the agent's state after a collision helps design more comprehensive reward functions and avoid such scenarios. Modeling and simulating collisions remain a significant challenge involving complex mechanics, geometry, and material science. However, we notice some interesting discussion on differentiable collision processes in recent research\cite{2023collision}.

The sim-to-real gap can be bridged by incorporating more detailed and diverse scenarios, realistic physics models, and accurate representations of sensor noise and failures. By creating more comprehensive and challenging simulations, developers can better prepare autonomous driving systems for the complexities of real-world operation. Meanwhile, through domain adaptation\cite{2024pamidomain_survey}, we can improve the transferability of models trained in simulated environments to real applications. It involves adjusting a model trained in the source domain (simulation) to perform well in the target domain (real world)\cite{2020domainadapt, 2023domainadapt}. This process aims to reduce the discrepancy between the two domains, allowing the model to generalize better to real-world conditions. Alternatively, we can integrate real-world data into the training and testing processes\cite{lan2023trafficgen, tan2021scenegen, scenarioNet} so that it is possible to refine and validate simulation models and ensure a more close matching to real conditions. 

\subsection{Safety Certificates}
Safety is a paramount concern in the development of AV. Ensuring the reliability and robustness of these systems involves addressing several critical aspects, from extensive real-world testing to mitigating issues related to multi-agent communication and data transfer.

\subsubsection{Safety in Real World} 
Industrial leaders emphasize the importance of cumulative test lengths to ensure the reliability of autonomous driving because real-world testing provides invaluable insights that cannot be fully captured in simulation. It allows for the evaluation of AVs in real traffic, weather, and road conditions and figures out the potential flaws that may not emerge during simulation. This type of testing is crucial for understanding how the system interacts with human drivers, pedestrians, and other road users, ensuring that it operates safely and effectively in a live environment. Although recent advancements\cite{qin2021learning, tro2021cbf, tro2023cbf} show the reliability of the hard safety constraints and the interpretation of barriers through learning methods, the long-tailed problem means we still cannot sample all risky scenarios and guarantee 100$\%$ safety with these approaches in the real world. Additionally, only a few research institutions and universities have access to specialized testing facilities for real-world traffic testing like MCity\cite{briefs2015mcity}. Although some studies include real-world validation and demonstrations, most are limited to down-scaled and simplified platforms\cite{2019primal, 2021primal2, GDQ_turtlebot3}. Consequently, the current safety validation of MARL-related research remains significantly inadequate. We believe this can be improved by accelerating the development of testing and research infrastructure and enhancing regional and international academic collaboration. 

\subsubsection{Private Information Concern}
MARL-based traffic systems rely heavily on data exchange and communication between agents. In MAS, AVs need to frequently share data related to their environment, position, and intended actions to coordinate effectively. This data exchange, while essential for the system's functioning, can expose sensitive information. Personal data, such as location history, can be vulnerable to unauthorized access if adequate security measures are not in place. External attacks significantly threaten the privacy and security of MARL systems in autonomous driving. Attackers can exploit vulnerabilities in AVs' communication protocols and data storage systems. To alleviate these concerns, developing robust communication protocols that include authentication and verification mechanisms can help prevent unauthorized access and data tampering. Meanwhile, real-time anomaly detection systems\cite{2021anomaly} can help identify and mitigate potential attacks by monitoring for unusual patterns or behaviors in the data exchange. Decentralized data storage\cite{sharma2018blockchain} can also enhance data security by providing a tamper-proof ledger of all data exchanges and transactions.

\subsection{Explainability}
Explainability emphasizes the understandable and causable decision-making process, which ensures the actions are transparent and reasonable. However, nowadays, most learning-based methods still adopt the black-box deep neural network as the main component. As the primary shortcomings of black-box models, the lack of transparency poses significant safety concerns\cite{ribeiro2016lime}. This opacity can lead to a lack of trust and confidence in the system, especially in safety-critical situations where understanding the reasoning behind a decision is crucial. For instance, if an AV makes an unexpected maneuver, it is essential to know whether the decision is based on valid reasoning or a flaw in the system.

Explainable methods like decision tree\cite{2021decisiontree} or rule-based systems\cite{2020rulebased} can be used to establish the mathematically understandable controller. However, fine system-level design needs expert domain knowledge. State representation learning can create a low-dimensional and meaningful representation of the state space by processing high-dimensional raw observation data\cite{2021explain_survey}. This approach captures the environmental variations influenced by the agent's actions, facilitating the extrapolation of explanations. Mainly, it is helpful in reinforcement learning for robotics and control, as it aids in understanding how the agent interprets observations and identifies what is relevant for learning to act effectively\cite{raffin2019srl1, traore2019srl2}. Besides, applying post-hoc analysis methods to existing models can help interpret their decisions\cite{krajna2022explain_survey2}. Techniques such as feature importance analysis, saliency map, and layer-wise relevance propagation can provide insights into which features and how these features influence specific decisions. Incorporating attention mechanisms in the model architecture can highlight which parts of the input data are most relevant to the decision-making process\cite{vaswani2017attention, kim2017explain_att, kim2018explain_att, 2022explain_att}. This can help in understanding the focus areas of the model during critical decision points. Another solution is using model-agnostic methods like local interpretable model-agnostic explanations~(LIME)\cite{ribeiro2016lime, zhao2021baylime} to provide the predictive explanations of any black-box model, which approximate the model's behavior locally to explain individual predictions.

\section{Future Directions}
\label{future}
In the last section, we briefly introduce the latest advancements and explain why we believe these directions are promising. We hope that this information will inspire researchers and lead to more outstanding research.

\subsection{Model-based MARL}
Model-based RL has achieved significant progress in single AV driving. By incorporating additional neural networks, we can model complex nonlinear dynamics and state transition functions\cite{hafner2019dreamer, hafner2020dreamerv2}. In recent research, researchers implement real-world high-speed racing in complicated tracks via extra 4 networks for environmental dynamics and prediction of future state, observation, and reward\cite{2022latent}. However, there is no free lunch. While model-based reinforcement learning offers better performance and improves explainability, it also increases the requirement for computational resources. Typically, centralized approaches model the environment through joint actions and observations and get rid of the non-stationarity and partial observability. However, scalability remains a significant challenge, especially for heterogeneous MAS. Conversely, decentralized approaches are easier to scale but struggle to reach consensus in non-stationary dynamics and partially observable environments. Beyond the design of communication protocols, recent research is exploring the possibility of abstracted and simplified modeling, such as inferring and predicting the subsequent actions of other agents. Additionally, decentralized paradigms introduce more networks for learning models, so both the selection of model representations and the design of efficient network architectures are promising topics.

\subsection{Development of Offline Multi-Agent Datasets}
\label{subsec::offline_data}
Offline paradigms can improve the practicality and realism of reinforcement learning. Specifically, online trial-and-error learning can cause financial losses and social disruption in mission-critical systems. Training policies in simulations suffer from an inherent gap between simulated environments and real dynamics and limits the ability to leverage extensive previously collected datasets\cite{offlinerl}. We observe that some recent studies are exploring how to use previously collected datasets to train single-vehicle offline reinforcement learning paradigms, paving the way for the expansion into MAS\cite{offline_ad}. As mentioned before, although independent learning can achieve optimal individual policies, it follows a self-centric optimization and lacks the validation of social preference and altruism. From this assumption, it is necessary to redevelop and collect datasets at the system level rather than the individual level. Combining with data augmentation approaches\cite{reuse2021data_aug, zhang2022data_aug, tits2023data_aug}, these datasets can be exponentially expanded to cover most of the daily and long-tailed scenarios. Additionally, there are opportunities to combine offline MARL with other emerging methods, such as safe-RL\cite{cmu2024safeoffline} or meta-RL\cite{offline_marl4ad}. In general, offline MARL and system-level datasets are essential for advancing large-scale autonomous driving. From pre-collected data and integrating simulations, we hope developers can create safer, more efficient, and more reliable autonomous driving systems in the future. 

\subsection{Human-in-the-Loop Learning}
Human-in-the-loop (HITL) learning\cite{wu2022hillearning} can enhance the effectiveness and safety of autonomous driving systems. These methods incorporate human feedback into the learning process, integrate human expertise and intuition, and ensure that systems can handle complex and dynamic real-world scenarios more effectively by leveraging human judgment to guide the learning process. Specifically, HITL learning involves human operators providing feedback on the system's performance during training. This feedback can come in various forms, such as corrections to the vehicle's actions, suggestions for alternative routes, or evaluations of the system's decision-making process. By integrating human feedback, autonomous driving systems can better identify and avoid potential hazards and reduce the probability of accidents. Besides, around human demonstrations, we can design extra modules to ensure the effective learning of policy correction and safety improvement\cite{wu2021hilad, peng2022HIP, li2022humanai}. Reinforcement Learning with Human Feedback (RLHF) is a recently proposed powerful paradigm that transforms human feedback into guidance signals. Particularly, it eliminates the requirement for manual reward design, which plays a vital role in robotic control and language models. Besides modeling the reward, RLHF can help autonomous systems quickly adapt to new and unexpected driving scenarios by incorporating real-time human feedback. In the latest research\cite{yuan2024unirlhf}, the first large-scale benchmark is proposed by integrating a series of RLHF baselines and over 1000 hybrid agents and expert driving trajectories in SMARTS\cite{zhou2020smarts} simulator. Meanwhile, its exciting results demonstrate that RLHF can follow the abstraction metrics and learn good human behavior. However, HITL means that it is inevitable that human driver participation will be required. How to minimize the need for human intervention, effectively extract key strategy improvements from these interventions, and prevent catastrophic forgetting will be important topics for future research in this field.

\subsection{Language Models for Autonomous Driving} 
As an eye-catching rising star, we have witnessed the contributions language models to the control communities\cite{chen2021decisiontrans, janner2021trajectorytrans, sun2024llm}. Although they are not necessarily RL frameworks, we still discuss their possibilities and future in our paper. Nowadays, large language models (LLMs)\cite{kenton2019bert, touvron2023llama, achiam2023gpt} and visual language models (VLMs)\cite{zhou2023vision, wen2024road} show significant potential for autonomous driving. These models can process and understand both textual and visual information and trigger new approaches to enhance the capabilities of AVs. LLMs can contribute to autonomous driving by providing sophisticated natural language understanding and generation support. These models can be used to interpret verbal instructions, understand traffic signs and signals, and process natural language inputs from passengers\cite{xu2023drivegpt4, sha2023languagempc, cui2024llm4ad}. Although they show expressive common sense reasoning and understanding capabilities, they are mostly leveraged only in simple decision-making due to the inability of visual information. VLMs combine visual and linguistic data to create a better understanding of the environment\cite{huang2023applications, zhou2023vision}. These models can interpret complex visual scenes and provide contextual understanding, which is crucial for making informed driving decisions. Moreover, they can be used to recognize and describe objects, predict the behavior of pedestrians and other vehicles, and understand the broader context of the driving environment\cite{wen2024road, li2024automated, pan2024vlp}. Recently, we discovered an interesting discussion and imagination of language models under the MARL paradigm in the latest review\cite{sun2024llm}. For instance, language models could simplify HITL interventions from complex manual operations to voice inputs. Alternatively, it can be used for context distillation\cite{snell2022learning} and make the complex observations and communication information in MAS more compact. Thus, the distilled features would be more suitable for edge computing platforms. This is a rapidly growing and highly significant field that warrants continuous attention.

\section*{Acknowledgement}
We thank Prof. Peng Yin from the City University of Hong Kong for sharing the structural design of Fig.~\ref{fig::history}. This work is supported by the European Union's Horizon 2020 Framework Programme for Research and Innovation under the Special Grant Agreement No. 945539 (Human Brain Project SGA3), in part by the National Natural Science Foundation of China (No. 62372329), in part by Shanghai Scientific Innovation Foundation (No.23DZ1203400), in part by Shanghai Rising Star Program (No.21QC1400900), in part by Tongji-Qomolo Autonomous Driving Commercial Vehicle Joint Lab Project, and in part by Xiaomi Young Talents Program.

\bibliographystyle{ieeetr}
\bibliography{main}

\vfill

\end{document}